\documentclass{article}

\usepackage{microtype}
\usepackage{graphicx}
\usepackage{subcaption}
\usepackage{booktabs} % for professional tables
\usepackage{multirow}
\usepackage{array}         % p{} 列格式，自定义列格式
\usepackage{colortbl}      % 支持 \rowcolor
\usepackage[table]{xcolor} % 支持表格着色和自定义颜色

\usepackage{hyperref}

\usepackage[preprint]{icml2026}

\usepackage{amsmath}
\usepackage{amssymb}
\usepackage{mathtools}
\usepackage{amsthm}
\usepackage{enumitem}
\usepackage[breakable,skins]{tcolorbox}
\tcbuselibrary{listings,breakable}

\usepackage{algorithm}



\usepackage{algpseudocode}
\usepackage[capitalize,noabbrev]{cleveref}

\theoremstyle{plain}

\theoremstyle{definition}

\theoremstyle{remark}

\usepackage[textsize=tiny]{todonotes}

\icmltitlerunning{On-Policy Supervised Fine-Tuning for Efficient Reasoning}

\begin{document}

\twocolumn[
  \icmltitle{On-Policy Supervised Fine-Tuning for Efficient Reasoning}



  \icmlsetsymbol{equal}{*}

  \begin{icmlauthorlist}
    \icmlauthor{Anhao Zhao}{EIT,polyu}
    \icmlauthor{Ziyang Chen}{EIT,PSL}
    \icmlauthor{Junlong Tong}{EIT,sjtu}
    \icmlauthor{Yingqi Fan}{EIT}
    \icmlauthor{Fanghua Ye}{tencent}
    \icmlauthor{Shuhao Li}{EIT,polyu}
    \icmlauthor{Yunpu Ma}{munich}
    \icmlauthor{Wenjie Li}{polyu}
    \icmlauthor{Xiaoyu Shen}{EIT}
  \end{icmlauthorlist}

  \icmlaffiliation{polyu}{Department of Computing, The Hong Kong Polytechnic University}
  \icmlaffiliation{PSL}{Université Paris Dauphine - PSL}
  \icmlaffiliation{EIT}{Eastern Institute of Technology, Ningbo}
  \icmlaffiliation{sjtu}{Shanghai Jiao Tong University}
  \icmlaffiliation{munich}{Ludwig Maximilian University of Munich}
  \icmlaffiliation{tencent}{Tencent Hunyuan / AI Lab}

  \icmlcorrespondingauthor{Xiaoyu Shen}{xyshen@eitech.edu.cn}

  \icmlkeywords{Machine Learning, ICML}

  \vskip 0.3in
]



\printAffiliationsAndNotice{}  % no special notice (required even if empty)

\begin{abstract}
Large reasoning models (LRMs) are commonly trained with reinforcement learning (RL) to explore long chain-of-thought reasoning, achieving strong performance at high computational cost.  Recent methods add multi-reward objectives to jointly optimize correctness and brevity, but these complex extensions often destabilize training and yield suboptimal trade-offs.
We revisit this objective and challenge the necessity of such complexity.
Through principled analysis, we identify fundamental misalignments in this paradigm: KL regularization loses its intended role when correctness and length are directly verifiable, and group-wise normalization becomes ambiguous under multiple reward signals. By removing these two items and simplifying the reward to a truncation-based length penalty, we show that the optimization problem reduces to supervised fine-tuning on self-generated data filtered for both correctness and conciseness. We term this simplified training strategy \textbf{on-policy SFT}.
Despite its simplicity, on-policy SFT consistently defines the accuracy–efficiency Pareto frontier. It reduces CoT length by up to \textbf{80}\% while maintaining original accuracy, surpassing more complex RL-based methods across five benchmarks. Furthermore, it significantly enhances training efficiency, reducing GPU memory usage by \textbf{50}\% and accelerating convergence by \textbf{70}\%. Our code is available at \url{https://github.com/EIT-NLP/On-Policy-SFT}.
%Overall, our results suggest that efficient reasoning benefits more from principled simplification than from increasingly complex RL objectives.
\end{abstract}

\section{Introduction}

The emergence of large reasoning models (LRMs), trained with reinforcement learning (RL) methods such as Group Relative Policy Optimization (GRPO)~\cite{deepseekmath} to explore reasoning trajectories, has led to strong performance across many tasks~\cite{cot,openai,LRM_survey,DeepSeek-R1,qwen3_technical_report}.
However, their performance relies on generating long chains of intermediate reasoning tokens, which incurs substantial computational and memory overhead~\cite{chen2025unveiling,ding2026llms}.
\begin{figure}[t]
  \centering
  \includegraphics[width=\linewidth]{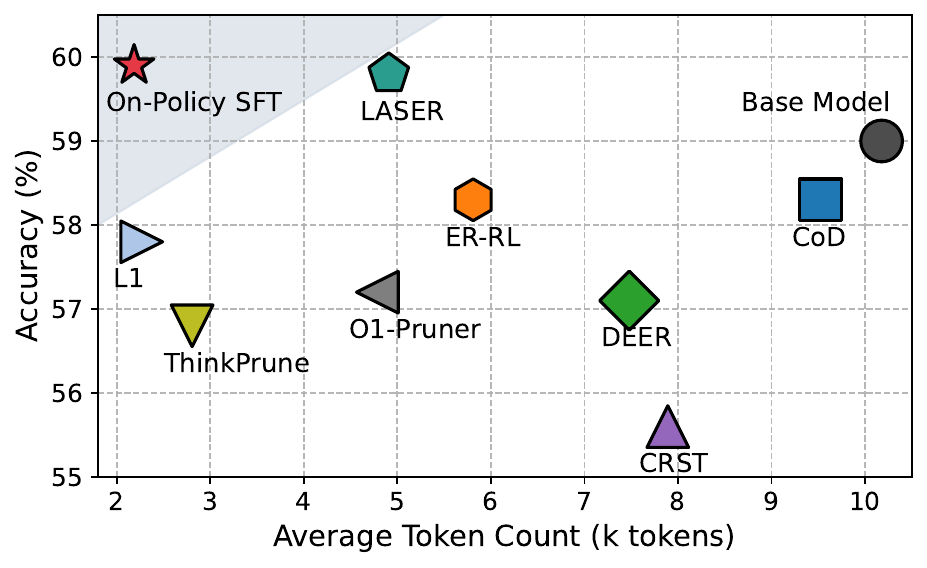}
  \vspace{-20pt}
  \caption{On-Policy SFT achieves a state-of-the-art accuracy–length trade-off on DeepSeek-R1-1.5B, reducing CoT length by approximately 80\% while slightly improving accuracy.}
  \label{fig:main_plot}
\end{figure}

To address this bottleneck, a growing body of work has explored efficient reasoning methods to shorten CoT while preserving accuracy~\cite{overthinking,thinkprune,efficient_reasoning_survey}. A common and intuitive strategy is to inherit the GRPO-based RL framework and reformulate efficient reasoning as a multi-reward optimization problem, in which auxiliary length-related rewards are introduced to encourage concise reasoning~\citep{L1,laser}.
Despite their conceptual appeal, such multi-reward formulations substantially increase reward-shaping complexity. In practice, they often lead to degraded accuracy, training instability, slower convergence, and heightened sensitivity to hyperparameter choices~\citep{dler_nvidia,xiaomi,er_rl,aware_first}. 

In this paper, we critically examine whether the indiscriminate adoption of complex objectives like GRPO aligns with the intrinsic structure of efficient reasoning. Through a principled analysis, we identify two fundamental misalignments in the current paradigm.
First, \emph{the KL-divergence regularization becomes redundant}. While essential in reinforcement learning from human feedback (RLHF) to mitigate reward overoptimization against noisy or learned reward models~\citep{rlhf_distribution_1,rlhf_distribution_2,rlhf}, it is unnecessary for common efficient reasoning tasks, where correctness and length are directly verifiable and free from distributional ambiguity.
Second, \emph{group-wise reward normalization becomes ill-suited}. Originally designed to stabilize training with single-reward signals~\citep{deepseekmath}, this mechanism can introduce optimization bias by amplifying gradients from uninformative samples~\citep{dr.grpo} and obscuring distinctions between candidate solutions in multi-reward scenarios, thereby introducing ambiguity into the learning signal~\cite{Group-Relative_Advantage_Biased}.
Furthermore, guided by recent findings that optimization objective design outweighs sophisticated reward shaping~\cite{dler_nvidia,justrl}, we adopt the simplest length penalty truncation, which assigns zero reward to responses exceeding a fixed length limit.
Taken together, by \textit{removing KL regularization, group-wise normalization, and adopting truncation-based rewards}, the original policy gradient objective reduces into a reward-free form equivalent to maximum-likelihood supervised fine-tuning (SFT).
Unlike standard SFT on fixed datasets, however, this method updates gradients using on-policy responses filtered for correctness and conciseness. We term this streamlined, reward-free training recipe \textbf{on-policy SFT}.

Despite its simplicity and without relying on reward shaping, on-policy SFT achieves state-of-the-art accuracy–length trade-offs across comparisons with \textbf{ten} strong baselines spanning training-free, SFT-based, and RL-based approaches, when evaluated on five widely used mathematics benchmarks.
On DeepSeek-R1-Distill-Qwen-1.5B, on-policy SFT attains an overall accuracy of 59.9\% and a Pass@\(N\) of 73.6\%, slightly outperforming the original model’s 59.0\% and 73.5\%, respectively, while reducing the average generation length from 10,178 tokens to 2,186 tokens, which corresponds to an approximate \textbf{80}\% reduction (see \autoref{fig:main_plot}).
When measured using \texttt{Eff}, a quantitative metric for accuracy–efficiency trade-offs, on-policy SFT yields a score of 2.74\%, surpassing the strongest RL-based baseline at 2.55\%. 
On DeepSeek-R1-Distill-Qwen-7B, on-policy SFT similarly achieves the highest \texttt{Eff} score of 2.97\%, while reducing generation length by roughly \textbf{70}\% relative to the original model. 
Furthermore, under varying generation-length budgets, on-policy SFT consistently lies on the accuracy–efficiency Pareto frontier.

Beyond inference-time efficiency, on-policy SFT also substantially improves training efficiency, reducing GPU memory consumption and wall-clock time per training step by approximately \textbf{50}\% across rollout configurations, and accelerating convergence by \textbf{70}\% compared to RL-based methods.
Moreover, on-policy SFT demonstrates \textbf{stronger length control} than RL-based baselines, as evidenced by lower generation length variance across multiple generations for the same input.
To better understand the source of these overall gains, we examine the role of the training data distribution. Specifically, we show that the effectiveness of on-policy SFT is primarily driven by the use of on-policy data.
Finally, we further present a set of empirically grounded training guidelines for on-policy SFT, including rollout temperature, the number of rollouts per input, length bias correction, and maximum output length. Together, these guidelines enable stable optimization and strong practical performance, and are accompanied by principled explanations of their underlying mechanisms.

Overall, our results suggest that some complex designs in existing approaches may be misaligned with efficient reasoning, and that simpler, principled alternatives can already achieve strong accuracy–efficiency Pareto frontiers at substantially lower training cost. More broadly, these findings indicate that future progress in efficient reasoning may benefit from prioritizing simplicity and principled design over increasing algorithmic complexity.
\section{Preliminary}
\label{Preliminary}
Given an question $q$, an LLM generates an output
$o = \{S, a\}$, where $S = (s_1, s_2, \ldots, s_T)$ denotes a sequence
of reasoning steps and $a$ is the final answer.
The objective of efficient reasoning is to learn a policy that minimizes reasoning length while preserving answer accuracy.

\paragraph{GRPO for Efficient Reasoning}
\label{GRPO for Efficient Reasoning}
GRPO has become a widely adopted algorithm for improving mathematical reasoning, driven by its success in training DeepSeek-R1~\cite{DeepSeek-R1}. 
Compared to PPO \citep{ppo}, GRPO removes the explicit value function and instead estimates advantages in a group-relative manner. Specifically, for a given question--answer pair $(q, a)$, the behavior policy $\pi_{\theta_{\text{old}}}$ samples a group of $G$ candidate responses $\{o_i\}_{i=1}^{G}$. The advantage of the $i$-th response is computed by normalizing the group-level rewards $\{R_i\}_{i=1}^{G}$:
\begin{equation}
\hat{A}_{i,t} =
\frac{R_i - \operatorname{mean}(\{R_i\}_{i=1}^{G})}
{\operatorname{std}(\{R_i\}_{i=1}^{G})}.
\label{eq:grpo_advantage}
\end{equation}
Similar to PPO, GRPO employs a clipped surrogate objective and additionally incorporates an explicit Kullback–Leibler (KL) divergence penalty:
\begin{equation}
\begin{aligned}
&J_{\text{GRPO}}(\theta)
=
\mathbb{E}_{(q,a)\sim\mathcal{D},\,\{o_i\}_{i=1}^{G}\sim\pi_{\theta_{\text{old}}}(\cdot\mid q)}
\\&\Bigg[
\frac{1}{G}\sum_{i=1}^{G}\frac{1}{|o_i|}
\sum_{t=1}^{|o_i|}
\min\Big(
r_{i,t}(\theta)\hat{A}_{i,t},\,
\operatorname{clip}\big(r_{i,t}(\theta),\\&1-\varepsilon,1+\varepsilon\big)\hat{A}_{i,t}
\Big)
- \beta\,\mathrm{KL}(\pi_\theta \,\|\, \pi_{\text{ref}})
\Bigg].
\end{aligned}
\end{equation}
where
$
r_{i,t}(\theta)
=
\frac{\pi_\theta(o_{i,t}\mid q, o_{i,<t})}
{\pi_{\theta_{\text{old}}}(o_{i,t}\mid q, o_{i,<t})}.
$
In GRPO, although practical implementations may perform multiple gradient updates per training step, the samples are generated by a very recent version of the current policy $\pi_{\theta_{\text{old}}}(\cdot\mid q)$. Consequently, gradients and advantages are estimated using data generated by the current or a very recent policy, and GRPO is therefore categorized as an on-policy algorithm \citep{revisit_grpo}.
As a reward-based policy gradient method, GRPO can be naturally extended to tasks that could be well defined by explicit reward functions. Since reasoning efficiency can be readily quantified as a scalar reward (e.g., via length penalties), recent studies~\citep{thinkprune,L1,correct_concise_and_complete} incorporate length-penalty rewards into GRPO to improve reasoning efficiency.

\paragraph{Reward Shaping for Efficient Reasoning}
\label{Reward Shaping for Efficient Reasoning}
While existing RL-based efficient reasoning methods employ a wide range of reward designs, we show that most can be cast into a unified formalization:
\begin{equation}
R_{\text{Eff}}(o \mid q)
=
R_{\text{Acc}}(o \mid q)
+
\gamma(o \mid q)\, R_{\text{Len}}(o \mid q),
\label{reward_unified_formalization}
\end{equation}
where $R_{\text{Acc}}(o \mid q)$ denotes the accuracy reward,
$R_{\text{Len}}(o \mid q)$ represents a length-related reward that
encourages shorter reasoning trajectories,
and $\gamma(o \mid q)$ controls the trade-off between correctness and efficiency.
The landscape of reward shaping strategies has gradually shifted toward greater complexity,
largely driven by increasingly elaborate choices of $R_{\text{Len}}(o \mid q)$ and $\gamma(o \mid q)$.
For example, prior work conditions $\gamma(o \mid q)$ on answer correctness or question difficult to selectively activate $R_{\text{Len}}(o \mid q)$, and defines $R_{\text{Len}}(o \mid q)$ using group-level statistics such as the average, median, or maximum length \citep{er_rl,xiaomi,aware_first}. Details are summarized in Appendix~\ref{app:Unified Reward Formulations}.

\section{From GRPO to On-Policy SFT}
\label{On-Policy SFT: A Minimalist Training Recipe}
{
\setlength{\tabcolsep}{2.5pt}
\renewcommand{\arraystretch}{1.4}
\begin{table*}[!t]
\centering
\caption{Comparison of on-policy SFT and baseline methods across the GSM8K, MATH-500, AMC23, AIME24, and AIME25 datasets. \textbf{Acc} denotes accuracy, \textbf{P@N} denotes Pass@\(N\), \textbf{Tok} indicates the average number of generated tokens, and \textbf{CR} represents the compression rate. In the table, \textcolor{yellow!60!white}{yellow} denotes \textit{training-free} methods; \textcolor{green!60!white}{green} highlights \textit{SFT-based} methods; and \textcolor{red!60!white}{red} indicates \textit{RL-based} methods.
}
\scalebox{0.58}{
\begin{tabular}{@{}lccccccccccccccccccccccccccc@{}} 
\toprule
 \multirow{3}{*}{\textbf{Method}}
 & \multicolumn{4}{|c}{\textbf{GSM8K}}
 & \multicolumn{4}{|c}{\textbf{MATH-500}}
 & \multicolumn{4}{|c}{\textbf{AMC23}}
 & \multicolumn{4}{|c}{\textbf{AIME24}}
 & \multicolumn{4}{|c}{\textbf{AIME25}}
 & \multicolumn{5}{|c}{\textbf{Overall}} \\
   & {Acc$\uparrow$} & {P@N$\uparrow$} & {Tok$\downarrow$} & {CR$\downarrow$} 
   & {Acc$\uparrow$} & {P@N$\uparrow$} & {Tok$\downarrow$} & {CR$\downarrow$} 
   & {Acc$\uparrow$} & {P@N$\uparrow$} & {Tok$\downarrow$} & {CR$\downarrow$} 
   & {Acc$\uparrow$} & {P@N$\uparrow$} & {Tok$\downarrow$} & {CR$\downarrow$} 
   & {Acc$\uparrow$} & {P@N$\uparrow$} & {Tok$\downarrow$} & {CR$\downarrow$} 
   & {Acc$\uparrow$} & {P@N$\uparrow$} & {Tok$\downarrow$} & {CR$\downarrow$} &  {Eff$\uparrow$} \\ 
\hline

\multicolumn{26}{l}{{\cellcolor[rgb]{0.957,0.957,0.957}}\textit{\textbf{DeepSeek-R1-Distill-Qwen-1.5B}}} \\

% Prompt Based
\textit{COT} & 85.0 & 92.0 & 2,533 & 100.0\% & 86.0 & 93.4 & 5,420 & 100.0\% & 73.0 & 92.5 & 9,171 & 100.0\% & 25.3 & 50.0 & 17,418 & 100.0\% & 26.0 & 40.0 & 16,348 & 100.0\% & \multicolumn{1}{|l}{~~59.0} & 73.5 & 10,178 & 100.0\%  &0.58\\
\rowcolor{yellow!20}
\textit{CoD} & 84.8 & 92.4 & 1,717 & 67.7\% & 83.8 & 91.8 & 5,044 & 93.0\% & 71.5 & 87.5 & 8,518 & 92.8\% & 28.6 & 53.3 & 17,058 & 97.9\% & \underline{22.6} & \underline{33.3} & 15,297 & 93.5\% & \multicolumn{1}{|l}{~~58.3} & 71.7 & 9,527 & 93.6\%  &0.61\\
\rowcolor{yellow!20}
\textit{DEER} & 83.5 & 92.0 & 1,091 & 43.0\% & 83.9 & 92.6 & 3,131 & 57.7\% & 71.0 & \underline{92.5} & 6,908 & 75.3\% & 26.6 & 46.6 & 12,964 & 74.4\% & 20.6 & \underline{33.3} & 13,305 & 81.3\% &\multicolumn{1}{|l}{~~57.1} & 71.4 &7,480&  73.5\%  &0.76 \\
% \rowcolor{yellow!20}
% \textit{NoThinking} & 76.5 & 87.4 & 297 & 11.7\% & 72.6 & 87.2 & 1,372 & 25.3\% & 58.5 & 82.5 & 2,477 & 27.0\% & 18.0 & 43.3 & 7,213 & 41.4\% & 13.3 & 23.3 & 4,990 & 30.5\% &\multicolumn{1}{|l}{~~47.7} & 64.7 & 27.2\% & -\\

% SFT
\rowcolor{green!20}
\textit{CRST} & 78.7 & 89.8 & 1,075 & 42.4\% & 83.5 & \underline{92.7} & 3,874 & 71.5\% & 70.5 & \underline{92.5} & 7,410 & 80.8\% & 25.3 & 53.3 & 13,954 & 80.1\% & 20.0 & 30.0 & 13,147 & 80.4\% &\multicolumn{1}{|l}{~~55.6} & 71.7 & 7,892 &77.5\% &0.70\\
\rowcolor{green!20}
\textit{TokenSkip} & 52.6 & 52.6 & 12,715 & 501.9\% & 52.0 & 52.0 & 13,685 & 252.4\% & 32.5 & 32.5 & 17,452 & 190.3\% & 3.3 & 3.3 & 24,889 & 142.8\% & 3.3 & 3.3 & 28,148 & 172.1\% & \multicolumn{1}{|l}{~~28.7} & 28.7 & 19,378 &190.4\%  &0.15\\
\rowcolor{green!20}
\textit{StepEntropy} & 77.1 & 87.7 & \textbf{507} & \textbf{20.0\%} & 64.2 & 80.6 & 4,475 & 82.5\% & 47.5 & 77.5 & 9,237 & 100.7\% & 8.6 & 30.0 & 15,437 & 88.6\% & 6.6 & 23.3 & 14,438 & 88.3\% &\multicolumn{1}{|l}{~~40.8} & 59.8 & 8,819 & 86.6\% & 0.46\\

% RL
% \rowcolor{red!10}
% \textit{AutoThink-Stage1} & 84.9 & 92.3 & 1,103 & 43.5\% & 82.4 & 90.4 & 2,603 & 48.0\% & 73.0 & 95.0 & 4,192 & 45.7\% & 26.0 & 43.3 & 10,472 & 60.1\% & 20.0 & 30.0 & 10,299 & 63.0\% & \multicolumn{1}{|l}{~~57.2} & 70.2 & 52.0\% & -\\
% \rowcolor{red!10}
% \textit{AutoThink-Stage2} & 87.0 & 92.9 & 1,624 & 64.1\% & 86.5 & 92.6 & 3,297 & 60.8\% & 73.0 & 90.0& 5,706 & 62.2\% & 36.0 & 63.3 & 9,697 & 55.6\% & 23.3 & 36.6 & 9,874 & 60.4\% & \multicolumn{1}{|l}{~~61.1} & 75.0 & 60.6\% & -\\
% \rowcolor{red!10}
% \textit{AutoThink-Stage3} & 85.5 & 92.2 & 591 & 23.3\% & 84.8 & 91.2 & 1,783 & 32.9\% & 76.0 & 90.0 & 3,430 & 37.4\% & 31.3 & 53.3 & 7,529 & 43.2\% & 23.3 & 43.3 & 7,550 & 46.1\% & \multicolumn{1}{|l}{~~60.1} & 74.0 & 36.6\% & -\\
\rowcolor{red!10}
\textit{ThinkPrune} & 84.7 & 91.8  & 710 & 28.0\% & 84.2 & 92.4 & \underline{1,546} & \underline{28.5\%} & 72.0 & \textbf{95.0} & 2,672 & 29.1\% & 25.3 & 46.6 & 4,888 & 28.1\% & 18.0 & \underline{33.3} & 4,213 &  25.8\% &\multicolumn{1}{|l}{~~56.8} & 71.8 & 2,806 & 27.6\%  &2.02\\
\rowcolor{red!10}
\textit{O1-Pruner} & 85.2 & \underline{92.8} & 916 & 36.1\% & 83.0 & 90.6 & 2,383 & 43.9\% & 72.5 & 90.0 & 3,942 & 42.9\% & 27.3 & 53.3 & 8,164 & 46.8\% & 18.0 & \underline{33.3} & 8,528 & 52.1\% &\multicolumn{1}{|l}{~~57.2} & 72.0 &4,787 & 47.0\%  &1.19\\
\rowcolor{red!10}
\textit{L1} & \underline{86.0} & \textbf{93.1} & 1,607 & 63.4\% & \underline{85.3} & 91.6 & 1,922 & 35.4\% & \textbf{74.5} & 90.0 & \underline{2,170} & \underline{23.7\%} & 23.3 & 43.3 & \textbf{2,881} & \textbf{16.5\%} & 20.0 & 30.0 & \textbf{2,762} & \textbf{16.9\%} & \multicolumn{1}{|l}{~~57.8} & 69.6 & \underline{2,268} & \underline{22.3\%}  &\underline{2.55}\\
\rowcolor{red!10}
\textit{ER-RL} & 83.1 & 92.0 & 721 & 28.5\% & \underline{85.3} & \textbf{93.6} & 2,466 & 45.5\% & 72.4 & \underline{92.5} & 4,913 & 53.6\% & 27.3 & 50.0 & 11,584 & 66.5\%  & \textbf{23.3} & \underline{33.3} & 9,373 & 57.3\% &\multicolumn{1}{|l}{~~58.3} & 72.3 &5,811& 57.1\%  &1.00\\
\rowcolor{red!10}
\textit{LASER} & \underline{86.0} & 92.2 & 1,178 & 46.5\% & \underline{85.3}  & 91.8 & 2,714 & 50.0\% & 72.0 & 90.0 & 4,262 & 46.4\% & \textbf{33.3} & \textbf{66.6} & 7,821 & 44.9\% & \underline{22.6} & \underline{33.3} & 8,569 & 52.4\% &\multicolumn{1}{|l}{~~\underline{59.8}}  & \textbf{74.8} & 4,909& 48.2\% &1.22\\
\textit{On-Policy SFT} & \textbf{86.1} & 92.1 & \underline{696} & \underline{27.5\%} & \textbf{85.7} & 92.5 & 1\textbf{,354} & \textbf{25.0\%} & \underline{73.5} & 90.0 & \textbf{1,977} & \textbf{21.6\%} & \underline{30.7} & \underline{56.7} & \underline{3,808} & \underline{21.8\%} & \textbf{23.3} & \textbf{36.7} & \underline{3,096} & \underline{18.9\%} & \multicolumn{1}{|l}{~~\textbf{59.9}} & \underline{73.6} & \textbf{2,186} &\textbf{21.5\%}  &\textbf{2.74}\\

\hline

\multicolumn{26}{l}{{\cellcolor[rgb]{0.957,0.957,0.957}}\textit{\textbf{DeepSeek-R1-Distill-Qwen-7B}}} \\
% Prompt Based
\textit{COT} & 92.5 & 95.3 & 1,701 & 100.0\% & 93.0 & 96.2 & 4,132 & 100.0\% & 88.5 & 95.0 & 6,722 & 100.0\% & 54.0 & 80.0 & 13,960 & 100.0\% & 38.6 & 56.6 & 14,478 & 100.0\% & \multicolumn{1}{|l}{~~73.3} & 84.6 & 8,199 & 100.0\% &0.89\\
\rowcolor{yellow!20}
\textit{CoD} & 91.4 & \underline{96.0} & 605 & 35.5\% & 91.6 & 95.6 & 2,694 & 65.2\% & \textbf{90.5} & 95.0 & 5,018 & 74.6\% & 51.3 & \underline{76.6} & 13,372 & 95.7\% & 38.0 & 56.6 & 14,202 & 98.0\% & \multicolumn{1}{|l}{~~\underline{72.6}} & 84.0 & 7,178 & 87.5\% &1.01\\
\rowcolor{yellow!20}
\textit{DEER} & 91.9 & 95.8 & 788 & 46.3\% & 90.3 & 95.8 & 2,508 & 60.7\% & \textbf{90.5} & 95.0 & 4,866 & 72.3\% & \underline{52.6} & \textbf{80.0} & 11,294 & 80.9\% & 36.6 & 56.6 & 11,786 & 81.4\% & \multicolumn{1}{|l}{~~72.4} & 84.6 & 6,248 & 76.2\% &1.16\\
% \rowcolor{yellow!20}
% \textit{NoThinking} & 86.7 & 95.1 & 253 & 14.8\% & 80.2 & 91.2 & 764 & 18.4\% & 61.0 & 85.0 & 1,415 & 21.0\% & 24.6 & 56.6 & 3,744 & 26.8\% & 16.6 & 26.6 & 2,611 & 18.0\% & \multicolumn{1}{|l}{~~53.8} & 70.9 & 19.8\% & -\\

% SFT
\rowcolor{green!20}
\textit{CRST} & 91.6 & 95.1 & 1,601 & 94.1\% & 91.3 & 95.6 & 3,879 & 93.8\% & \underline{88.5} & 92.5 & 6,337 & 94.2\% & 50.0 & \underline{76.6} & 13,841 & 99.1\% & 36.6 & 53.3 & 14,785 & 102.0\% & \multicolumn{1}{|l}{~~71.6} & 82.6 & 8,089 & 98.7\% &0.89\\
\rowcolor{green!20}
\textit{TokenSkip} & 86.1 & 86.1 & 4,323 & 254.1\% & 79.6 & 79.6 & 6,584 & 159.3\% & 72.5 & 72.5 & 9,459 & 140.7\% & 23.3 & 23.3 & 22,443 & 160.7\% & 16.6 & 16.6 & 21,383 & 147.6\% & \multicolumn{1}{|l}{~~55.6} & 55.6 & 12,838 & 156.6\% &0.43\\
\rowcolor{green!20}
\textit{StepEntropy} & 87.6 & 93.6 & \underline{487} & \underline{28.6\%} & 71.2 & 85.2 & 4,914 & 118.9\% & 48.5 & 77.5 & 9,734 & 144.8\% & 10.0 & 26.6 & 15,986 & 114.5\% & 12.6 & 20.0 & 15,618 & 107.8\% & \multicolumn{1}{|l}{~~46.0} & 60.6 & 9,348 & 114.0\% &0.49\\

% % RL
% \rowcolor{red!10}
% \textit{AutoThink-Stage1} & 91.7 & 95.3 & 564 & 33.1\% & 90.3 & 95.2 & 1,637 & 39.6\% & 84.5 & 95.0 & 3,768 & 56.0\% & 47.3 & 70.0 & 8,318 & 59.5\% & 35.3 & 50.0 & 9,437 & 65.1\% & \multicolumn{1}{|l}{~~69.8} & 81.1 & 50.7\% & -\\
% \rowcolor{red!10}
% \textit{AutoThink-Stage2} & 92.4 & 95.5 & 823 & 48.3\% & 91.6 & 95.6 & 2,336 & 56.5\% & 91.0 & 95.0 & 3,834 & 57.0\% & 57.3 & 83.3 & 7,342 & 52.5\% & 40.0 & 56.6 & 8,817 & 60.9\% & \multicolumn{1}{|l}{~~74.4} & 85.2 & 55.0\% & -\\
% \rowcolor{red!10}
% \textit{AutoThink-Stage3} & 92.2 & 96.5 & 691 & 40.6\% & 91.2 & 95.0 & 2,189 & 52.9\% & 92.0 & 95.0 & 3,623 & 53.9\% & 52.6 & 73.3 & 7,481 & 53.5\% & 38.6 & 60.0 & 9,061 & 62.5\% & \multicolumn{1}{|l}{~~73.3} & 83.9 & 52.7\% & -\\
\rowcolor{red!10}
\textit{ThinkPrune} & \underline{92.2} & 95.7 & 862 & 50.7\% & 91.3 & 95.2 & 1,984 & 48.0\% & 88.0 & 95.0 & 3,397 & 50.5\% & 45.9 & 70.0 & 7,181 & 51.4\% & 31.3 & 43.3 & 7,514 & 51.9\% & \multicolumn{1}{|l}{~~69.7} & 79.8 & 4,188 & 51.1\% &1.67\\
\rowcolor{red!10}
\textit{O1-Pruner} & 92.0 & \textbf{96.2} & 1,439 & 84.6\% & \textbf{92.4} & 96.0 & 3,617 & 87.5\% & \textbf{90.5} & 95.0 & 5,379 & 80.0\% & 50.0 & \textbf{80.0} & 12,791 & 91.6\% & \underline{39.3} & \textbf{63.3} & 12,585 & 86.9\% & \multicolumn{1}{|l}{~~\textbf{72.8}} & \textbf{86.1} & 7,162 & 87.4\% &1.02\\
\rowcolor{red!10}
\textit{L1} & 91.7 & 94.9 & 1,371 & 80.6\% & 91.4 & 95.3 & \underline{1,708} & \underline{41.3\%} & 83.5 & 92.5 & \underline{2,191} & \underline{32.6\%} & 39.3 & \underline{76.6} & \textbf{3,390} & \textbf{24.3\%} & 30.6 & 53.3 & \textbf{3,122} & \textbf{21.6\%} & \multicolumn{1}{|l}{~~67.3} & 82.5 & \textbf{2,356} & \textbf{28.7\%} &\underline{2.86}\\
\rowcolor{red!10}
\textit{ER-RL} & 87.3 & 94.1 & \textbf{371} & \textbf{21.8\%} & 90.9 & \underline{96.1} & 2,014 & 48.7\% & 87.5 & 95.0 & 4,221 & 62.8\% & \textbf{53.3} & \textbf{80.0} & 9,858 & 70.6\% & \underline{39.3} & 56.6 & 10,895 & 75.3\% & \multicolumn{1}{|l}{~~71.7} & 84.4 & 5,472 & 66.7\% &1.31\\
\rowcolor{red!10}
\textit{LASER} & 92.1 & 95.9 & 930 & 54.6\% & \underline{92.0} & \textbf{96.4} & 1,857 & 44.9\% & \underline{88.5} & \underline{95.5} & 3,137 & 46.6\% & 51.3 & 73.3 & 6,104 & 43.7\% & \textbf{40.0} & \underline{60.0} & 6,270 & 43.3\% &\multicolumn{1}{|l}{~~\textbf{72.8}} & 84.2 & 3,660 & 44.6\% &1.99\\
\textit{On-Policy SFT} & \textbf{92.6} & \underline{96.0} & 498 & 29.3\% & 91.4 & 95.0 & \textbf{1,178} & \textbf{28.5\%} & 87.5 & \textbf{97.5} & \textbf{1,961} & \textbf{29.2\%} & \underline{52.6} & \textbf{80.0} & \underline{4,594} & \underline{32.9\%} & \textbf{40.0} & \underline{60.0} & \underline{4,039} & \underline{27.9\%} & \multicolumn{1}{|l}{~~\textbf{72.8}} & \underline{85.7} & \underline{2,454} & \underline{29.9\%} &\textbf{2.97}\\

\hline

\end{tabular}
}
% \vspace{3pt}
\label{main_table}
\end{table*}
}

Although existing RL-based efficient reasoning methods substantially reduce CoT length, they often incur accuracy degradation that varies across tasks of different complexity \citep{dler_nvidia,correct_concise_and_complete}. These suboptimal accuracy–efficiency trade-offs motivate us to re-examine whether such indiscriminate inheritance of complex optimization objectives is well aligned with the intrinsic properties of the efficient reasoning problem.

\subsection{Revisiting the GRPO Objective}
\label{Revisiting the GRPO Optimization Objective}
We ground our analysis in two distinctive properties of the efficient reasoning setting: \textit{directly verifiable correctness and length penalties}, and \textit{its inherently multi-reward nature}. Based on these characteristics, we examine different components in GRPO and identify potential mismatches.

\paragraph{KL Divergence}
The KL divergence term is commonly adopted in reinforcement learning
from human feedback (RLHF)~\citep{rlhf},
where the reward is provided by a learned reward model trained on data
collected under a reference policy $\pi_{\text{ref}}$.
In this setting, KL regularization plays a critical role in preventing
the optimized policy $\pi_\theta$ from deviating excessively from the distribution
over which the reward model is reliable
\citep{rlhf_distribution_1,rlhf_distribution_2,dr.grpo}.
In contrast, for efficient reasoning tasks, answer accuracy is verifiable, and length penalties can be quantified as scalar rewards, thereby eliminating the need for a learned reward model.
As a result, concerns about reward unreliability under distributional shift
do not arise.
This suggests that \emph{the KL divergence term may be unnecessary
and may even overly constrain policy updates during optimization}.
Removing this component can not only simplify the training objective, but also reduce the memory and computational overhead associated with maintaining the reference policy $\pi_{\text{ref}}$ during training.

\paragraph{Group-wise Reward Normalization}
GRPO estimates advantages by normalizing rewards within each rollout group, thereby avoiding explicit value function approximation (as in PPO) and stabilizing training~\citep{deepseekmath}. However, despite its effectiveness in incentivizing reasoning capability, this normalization mechanism may be ill-suited to the efficient reasoning setting.
First, \textit{question-level normalization results in varying weights across different questions.}
When rewards within a group vary little (e.g., all responses are correct and similarly long), the resulting small standard deviation in the denominator can amplify gradient magnitudes.
In practice, such samples provide limited learning signal yet are overemphasized during optimization, potentially leading to training instability~\citep{dr.grpo,Group-Relative_Advantage_Biased}.
Second, \textit{GRPO is fundamentally designed to optimize a single scalar reward objective, whereas efficient reasoning operates in a multi-reward regime}. 
This mismatch can introduce ambiguity in the learning signal when heterogeneous reward components are aggregated and normalized~\citep{gdpo}.
For instance, consider two rollouts with reward vectors \((0, 1)\) and \((0, 0)\), which yield scalar rewards of \(1\) and \(0\) under summation. After group-wise normalization, the resulting advantages are \((-0.7071,\, 0.7071)\). Notably, the same normalized advantages also arise for reward vectors \((1, 1)\) and \((0, 0)\), despite their fundamentally different reward compositions.
This observation motivates us to \emph{forego group-wise reward normalization and instead optimize rewards directly using a baseline}.

% \paragraph{The Simplest Length Penalty: Truncation}
% \label{The Simplest Length Penalty: Truncation}
% Prior work on RL-based efficient reasoning frequently reports improved accuracy–length trade-offs through increasingly sophisticated reward shaping \citep{aware_first,laser,xiaomi}. However, a fundamental question remains insufficiently examined: \textit{are these improvements intrinsically driven by the necessity of complex reward formulations, or do they primarily compensate for misaligned optimization objectives?}
% To investigate this question, we pair the efficient reasoning-aligned objective with a deliberately minimalist length penalty—truncation \citep{thinkprune,dler_nvidia}.
% Specifically, truncation defines a binary reward: a response receives a value of 1 if and only if it is correct and falls within a predefined length threshold (typically a few thousand tokens), and 0 otherwise.
% Under this configuration, we evaluate whether the aligned optimization objective, combined with a minimal reward signal, is sufficient to achieve competitive accuracy–efficiency trade-offs. This controlled experimental setup allows us to assess whether prior reward-shaping complexity is fundamentally required by the task or instead serves primarily to offset optimization-level misalignment.
\paragraph{The Simplest Length Penalty: Truncation}
Recent evidence suggests that the principled design of the optimization objective plays a more critical role than sophisticated reward design \citep{dler_nvidia,justrl}. In particular, \citet{dler_nvidia} show that even the simplest truncation strategy—assigning zero reward to responses exceeding a fixed length threshold—can achieve state-of-the-art accuracy–efficiency trade-offs when combined with an appropriate RL optimization objective.
Motivated by this finding, we conduct a parallel investigation. Instead of introducing additional complexity into reward shaping, we adopt the same minimalist truncation reward and pair it with our revised optimization objective. This design enables us to isolate the role of the optimization formulation and assess whether strong performance can be attained without intricate reward engineering.
Formally, truncation defines a binary reward:
\begin{equation}
R_{\text{trunc}}(o \mid q) =
\begin{cases}
1, & \text{if $o$ is correct and $|o| \le \tau$,} \\
0, & \text{otherwise,}
\end{cases}
\end{equation}
where $\tau$ denotes a predefined length threshold. Compared to the general composite formulation in \autoref{reward_unified_formalization}, truncation removes auxiliary length components and trade-off coefficients, collapsing correctness and conciseness into a single unified criterion.

\subsection{On-Policy SFT}
\label{On-Policy SFT subsection}
\begin{figure*}[t]
    \centering
    \includegraphics[width=\linewidth]{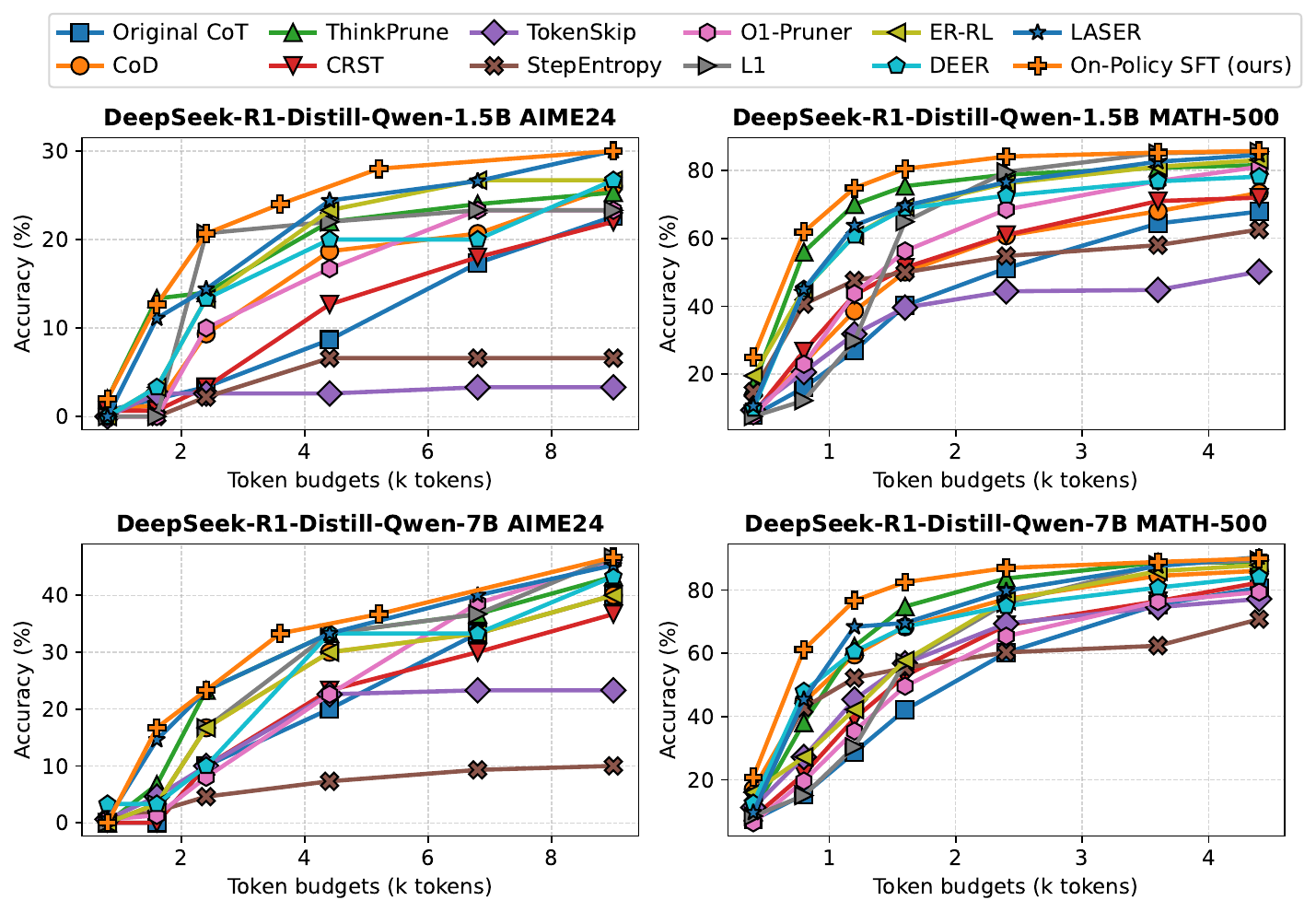}
    % \vspace{-19pt}
    \caption{Performance--efficiency trade-offs of on-policy SFT and baseline methods under varying generation token budgets.}
    \label{fig:acc_vs_tokens}
\end{figure*}
% In this subsection, we examine the optimization objective obtained by (i) simplifying the GRPO objective, specifically by removing the KL divergence term and eliminating the denominator corresponding to the standard deviation in group-wise reward normalization, and (ii) adopting a sparse binary truncation reward. Our analysis is conducted from a gradient-based perspective. We begin by considering the gradient of the original GRPO objective:
In this subsection, we analyze the optimization objective resulting from the analyses in \autoref{Revisiting the GRPO Optimization Objective} from a gradient-based perspective. We begin by considering the gradient of the original GRPO objective:
\begin{equation}
\begin{aligned}
&\nabla_\theta J_{\text{GRPO}}(\theta)
=
\mathbb{E}_{(q,a)\sim\mathcal{D},\,\{o_i\}_{i=1}^{G}\sim\pi_{\theta_{\text{old}}}(\cdot\mid q)}
\Bigg[
\\&\frac{1}{G}\sum_{i=1}^{G}\frac{1}{|o_i|}
\sum_{t=1}^{|o_i|}
\Big(
\hat{A}_{i,t}
+
\beta\Big(
\frac{\pi_{\text{ref}}(o_{i,t}\mid o_{i,<t})}
{\pi_\theta(o_{i,t}\mid o_{i,<t})}
- 1
\Big)
\Big)
\\&\nabla_\theta \log \pi_\theta(o_{i,t}\mid q, o_{i,<t})
\Bigg].
\end{aligned}
\end{equation}
% A detailed derivation is provided in Appendix~\ref{app:grpo_grad}. 
% First, following the analysis in \autoref{Revisiting the GRPO Optimization Objective}, we remove the KL divergence regularization term
% and replace group-wise reward normalization
% with the raw reward signal.
% The resulting gradient simplifies to:
% \begin{equation}
% \begin{aligned}
% &\nabla_\theta J(\theta)
% =
% \mathbb{E}_{(q,a)\sim\mathcal{D},\,\{o_i\}_{i=1}^{G}\sim\pi_{\theta_{\text{old}}}(\cdot\mid q)}\\
% &\Bigg[
% \frac{1}{G}\sum_{i=1}^{G}\frac{1}{|o_i|}
% \sum_{t=1}^{|o_i|}
% \hat{R}_{i,t}
% \nabla_\theta \log \pi_\theta(o_{i,t}\mid q, o_{i,<t})
% \Bigg].
% \end{aligned}
% \label{eq:policy-gradient}
% \end{equation}

A detailed derivation is provided in Appendix~\ref{app:grpo_grad}. We first simplify the GRPO objective by removing the KL divergence–related term
$\beta\!\left(\frac{\pi_{\text{ref}}(o_{i,t}\mid o_{i,<t})}{\pi_\theta(o_{i,t}\mid o_{i,<t})}-1\right)$
and by eliminating the standard-deviation $\operatorname{std}(\{R_i\}_{i=1}^{G})$ in reward normalization. The resulting policy gradient simplifies to:
\begin{equation}
\begin{aligned}
&\nabla_\theta J(\theta)
=
\mathbb{E}_{(q,a)\sim\mathcal{D},\,\{o_i\}_{i=1}^{G}\sim\pi_{\theta_{\text{old}}}(\cdot\mid q)}\\&
\Bigg[
\frac{1}{G}\sum_{i=1}^{G}\frac{1}{|o_i|}
\sum_{t=1}^{|o_i|}
\hat{R}_{i,t}\,
\nabla_\theta \log \pi_\theta(o_{i,t}\mid q, o_{i,<t})
\Bigg],
\end{aligned}
\label{eq:policy-gradient}
\end{equation}
where $\hat{R}_{i,t} = R_i - \operatorname{mean}(\{R_i\}_{i=1}^{G})$. The reward baseline $\operatorname{mean}(\{R_i\}_{i=1}^{G})$ reduces gradient variance and helps stabilize training in challenging regimes that require balancing exploration and exploitation, such as training models from weak to strong reasoning ability~\citep{DeepSeek-R1,advantage_importance}.
In our setting, however, we focus on models that already exhibit strong reasoning ability, where the objective is to favor shorter solutions among existing reasoning trajectories, emphasizing exploitation over exploration. Accordingly, we set $\hat{R}_{i,t} = R_i$, which further simplifies the derivation without affecting the intended optimization behavior.
Notably, after this simplification, we observe that the GRPO gradient reduces to a form that is mathematically equivalent to the classical \textit{REINFORCE}~\citep{REINFORCEMENT} policy gradient, as detailed in the Appendix~\ref{app:reinforce}.

In GRPO, the reward assigned to a rollout is shared across all tokens. Accordingly, for a fixed rollout $o_i$, the reward term is constant, i.e., $\hat{R}_{i,t} = \hat{R}_i$ for all $t = 1, \ldots, |o_i|$. We therefore drop the token index $t$ and express the reward as depending only on the rollout index $i$, yielding
\begin{equation}
\begin{aligned}
&\nabla_\theta J(\theta)
=
\mathbb{E}_{(q,a)\sim\mathcal{D},\,\{o_i\}_{i=1}^{G}\sim\pi_{\theta_{\text{old}}}(\cdot\mid q)}\\&
\Bigg[
\frac{1}{G}\sum_{i=1}^{G} \hat{R}_i\,\frac{1}{|o_i|}
\sum_{t=1}^{|o_i|}
\nabla_\theta \log \pi_\theta(o_{i,t}\mid q, o_{i,<t})
\Bigg].
\end{aligned}
\end{equation}

Next, following \autoref{Revisiting the GRPO Optimization Objective}, we instantiate the raw reward as a binary truncation reward. Specifically, $\hat{R}_i \in \{0,1\}$, where $\hat{R}_i = 1$ if the response is correct within a predefined fixed length limit $L$, and $\hat{R}_i = 0$ otherwise. We represent this reward using an indicator function $\mathbf{1}\{\hat{R}_i = 1\}$. Let $\mathcal{C}_L$ denote the set of responses for which $\hat{R}_i = 1$. The indicator can then be equivalently expressed as the set-based constraint $\mathbf{1}\{o_i \in \mathcal{C}_L\}$. Substituting this expression yields
\begin{equation}
\begin{aligned}
&\nabla_\theta J(\theta)
=
\mathbb{E}_{(q,a)\sim\mathcal{D},\,\{o_i\}_{i=1}^{G}\sim\pi_{\theta_{\text{old}}}(\cdot\mid q)}
\Bigg[
\frac{1}{G}\sum_{i=1}^{G}\\&
\mathbf{1}\{o_i \in \mathcal{C}_L\}\,
\frac{1}{|o_i|}
\sum_{t=1}^{|o_i|}
\nabla_\theta \log \pi_\theta(o_{i,t}\mid q, o_{i,<t})
\Bigg].
\end{aligned}
\label{gradient_of_on_policy_sft}
\end{equation}
Surprisingly, we observe that the \textit{resulting gradient shares the same functional
form as the gradient of the maximum likelihood objective} used in standard
supervised fine-tuning:
\begin{equation*}
\nabla_\theta J_{\text{SFT}}(\theta)
=
\mathbb{E}_{(q,o)\sim\mathcal{D}_{\text{SFT}}}
\Bigg[
\frac{1}{|o|}
\sum_{t=1}^{|o|}
\nabla_\theta \log \pi_\theta(o_t \mid q, o_{<t})
\Bigg].
\end{equation*}
The only difference lies in the data distribution.
Rather than optimizing over a fixed supervised dataset $\mathcal{D}_{\text{SFT}}$,
our updates are computed on responses generated on-policy by
$\pi_{\theta_{\text{old}}}$, where $\mathbf{1}\{o_i \in \mathcal{C}_L\}$ serves as a
selection mechanism enforcing correctness and length constraints.
% As a result, Optimizing the objective in \autoref{gradient_of_on_policy_sft} is equivalent to applying the standard SFT loss to on-policy samples, \textit{without relying on an explicit reward signal}.
% \begin{figure}[t]
%     \centering
%     \includegraphics[width=\linewidth]{Plot/GPU_Memory_vs_Batch_Size_1.5b_gpu.pdf}
%     \vspace{-19pt}
%     \caption{Average GPU memory consumption and wall-clock time per training step of on-policy SFT and the RL-based baseline ThinkPrune under varying rollout numbers.}
%     \label{fig:gpu_memory}
% \end{figure}
% \begin{figure}[t]
%     \centering
%     \includegraphics[width=\linewidth]{Plot/std_plot_1.5b.pdf}
%     \vspace{-19pt}
%     \caption{Length control comparison between on-policy SFT and RL-based baseline ThinkPrune measured by the coefficient of variation under varying token budgets on AIME24.
% }
%     \label{fig:std_plot}
% \end{figure}
% \begin{figure}[t]
%     \centering
%     \includegraphics[width=\linewidth]{Plot/off_policy_and_on_policy_7b.pdf}
%     \vspace{-16pt}
%     \caption{Comparison between on-policy SFT and its off-policy variant on MATH-500 for DeepSeek-R1-Qwen-7B-distilled.}
%     \label{fig:on_policy_and_off_policy}
% \end{figure}

Finally, we examine the dependence on response length $|o_i|$. Unlike pretraining \citep{pre_training}, where all tokens are packed into a fixed-length context and the loss is normalized by the context length, self-generated CoT responses vary substantially in length.
As a result, the $|o_i|$ term in \autoref{gradient_of_on_policy_sft} can induce disproportionately large gradients for shorter, potentially noisy responses, biasing training toward suboptimal solutions.
To mitigate this issue, we replace $|o_i|$ with the maximum response length within the batch, ensuring uniform gradient magnitudes across all tokens.
Rewriting the corrected gradients as a loss function yields our final optimization objective:
\begin{equation}
\begin{aligned}
&J(\theta)
=
\mathbb{E}_{(q,a)\sim\mathcal{D},\,\{o_i\}_{i=1}^{G}\sim\pi_{\theta_{\text{old}}}(\cdot\mid q)}
\Bigg[
\frac{1}{G}\sum_{i=1}^{G}\\&
\mathbf{1}\{o_i \in \mathcal{C}_L\}\,
\frac{1}{\max_j |o_j|}
\sum_{t=1}^{|o_i|}
\log \pi_\theta(o_{i,t}\mid q, o_{i,<t})
\Bigg] \\
&=
c_L\;
\underbrace{\mathbb{E}_{o_i \sim \mathcal{D}_L^{+}}
\Bigg[
\frac{1}{\max_j |o_j|}
\sum_{t=1}^{|o_i|}
\log \pi_\theta(o_{i,t}\mid q, o_{i,<t})
\Bigg]}_{J_{\text{SFT}}(\theta)},
\end{aligned}
\label{eq:on_policy_sft}
\end{equation}
where $\mathcal{D}_L^{+}=\{o_i \mid (q,a)\sim\mathcal{D},\,o_i\sim\pi_{\theta_{\text{old}}}(\cdot\mid q),\,o_i\in\mathcal{C}_L\}$, and
$c_L=\mathbb{E}_{(q,a)\sim\mathcal{D},\,o_i\sim\pi_{\theta_{\text{old}}}(\cdot\mid q)}[\mathbf{1}\{o_i\in\mathcal{C}_L\}]$ is a constant independent of $\theta$.
We adopt \autoref{eq:on_policy_sft} as the optimization objective and refer to the resulting minimalist training approach as \textbf{on-policy supervised fine-tuning (SFT)}, reflecting its use of the standard SFT loss under an on-policy data distribution. For completeness, we present the full training procedure of on-policy SFT in Algorithm~\ref{alg:onpolicy-sft}.

\section{Experimental Setup}
\label{Experimental Setup}

\begin{figure*}[t]
  \centering

  \begin{minipage}{0.48\textwidth}
    \centering
    \includegraphics[width=\linewidth]{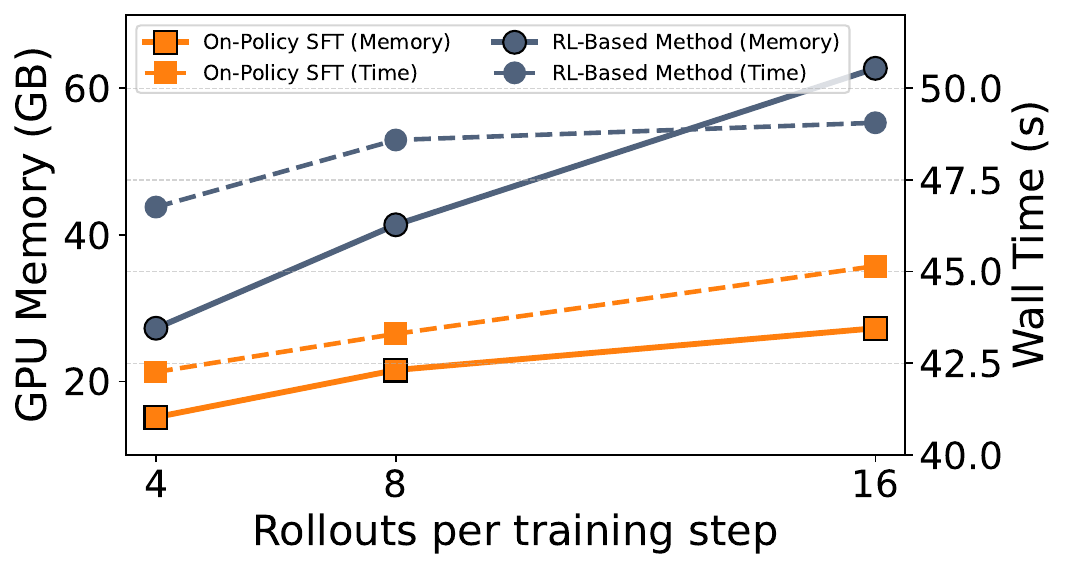}
    \vspace{-17pt}
    \caption{Average GPU memory consumption and wall-clock time per training step of on-policy SFT and the RL-based baseline ThinkPrune for the 1.5B model under varying rollout numbers.}
    \label{fig:gpu_memory}
  \end{minipage}\hfill
  \begin{minipage}{0.48\textwidth}
    \vspace{-3pt}
    \centering
    \includegraphics[width=\linewidth]{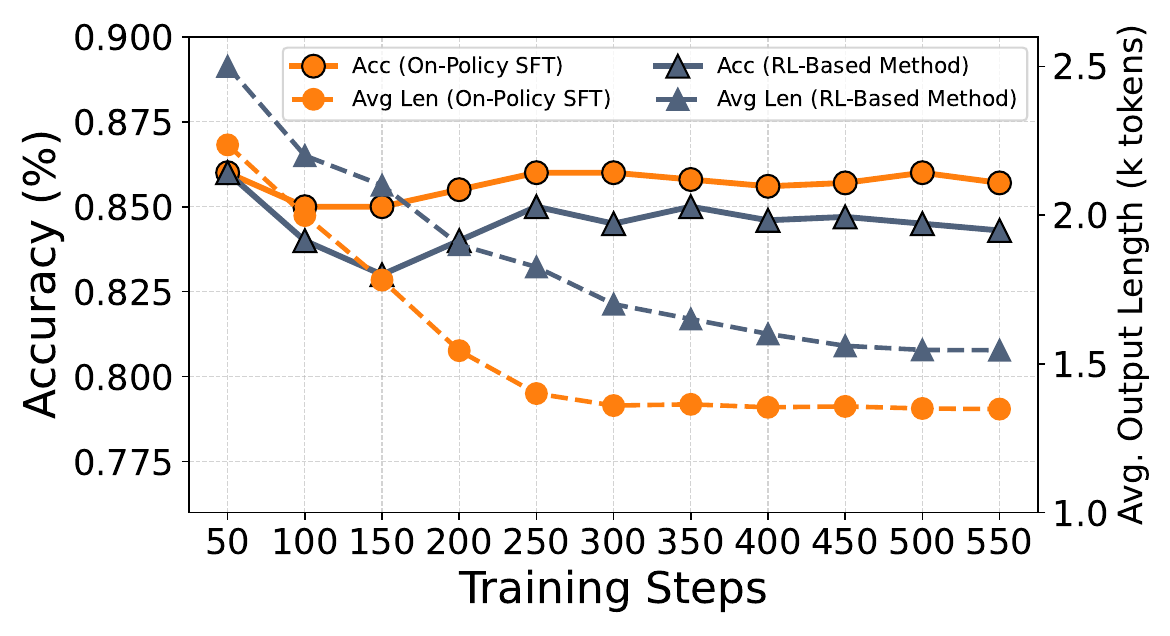}
    \vspace{-18pt}
    \caption{Convergence speed of on-policy SFT and RL-based baseline ThinkPrune on MATH-500 (1.5B; rollout = 8, batch size = 64).}
    \label{fig:on_policy_sft_vs_thinkprune}
  \end{minipage}

  \par\vspace{1em}  % 关键：换行
  \begin{minipage}{0.48\textwidth}
    \centering
    \includegraphics[width=\linewidth]{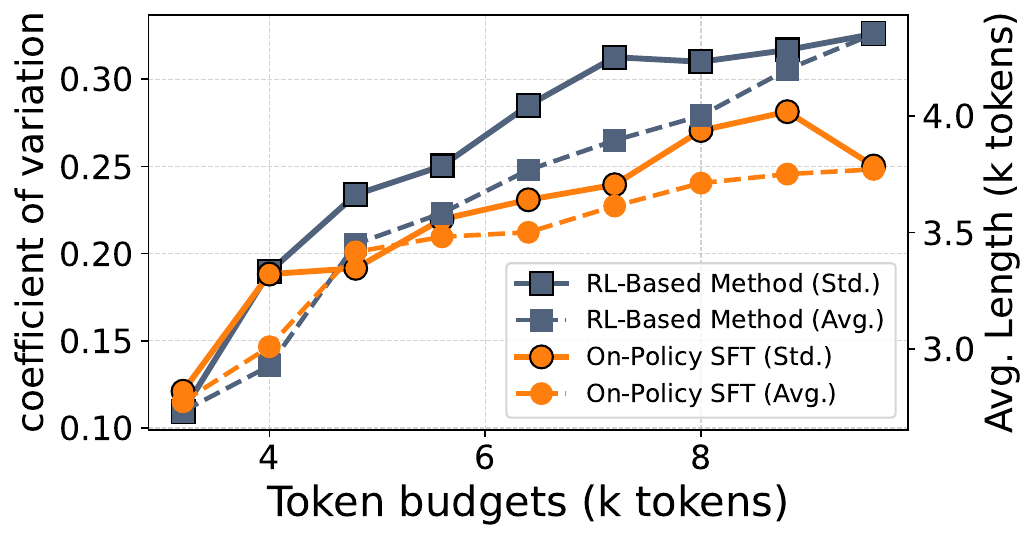}
    \vspace{-20pt}
    \caption{Length control comparison between on-policy SFT and RL-based baseline ThinkPrune measured by the coefficient of variation under varying token budgets on AIME24.}
    \label{fig:std_plot}
  \end{minipage}\hfill
  \begin{minipage}{0.48\textwidth}
    \vspace{-17pt}
    \centering
    \includegraphics[width=\linewidth]{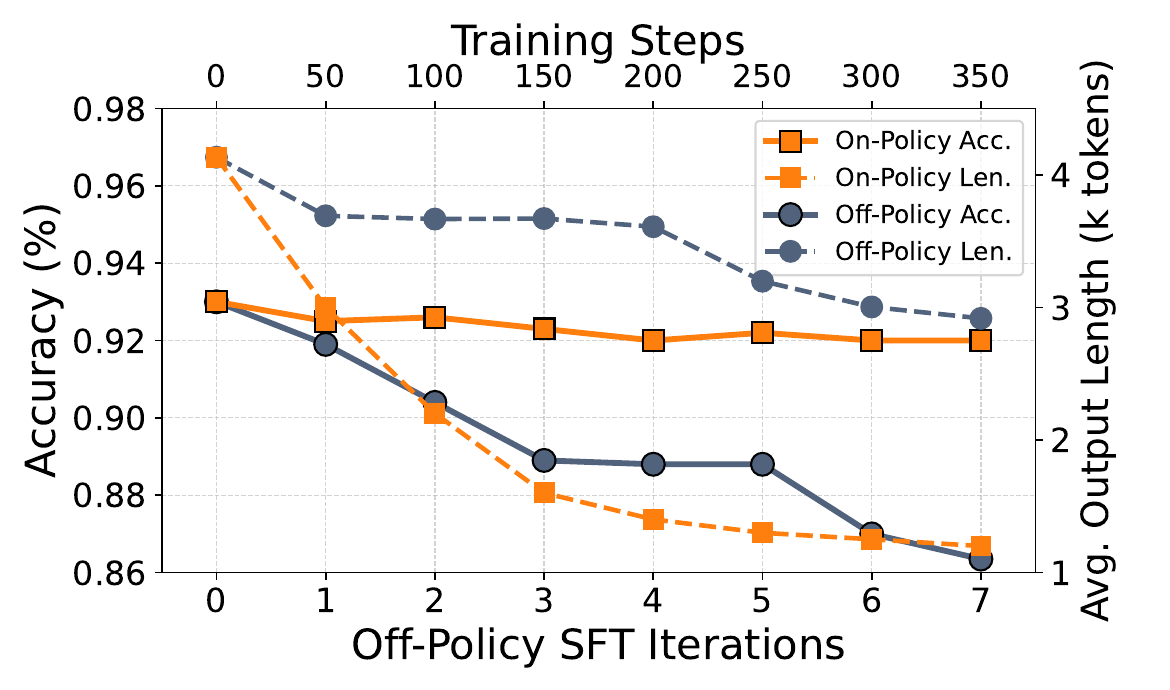}
    \vspace{-22pt}
    \caption{Comparison between on-policy SFT and its off-policy variant on MATH-500 for DeepSeek-R1-Qwen-7B-distilled.}
    \label{fig:on_policy_and_off_policy}
  \end{minipage}\hfill

\end{figure*}
\paragraph{Models and Training Dataset}
We consider representative open-sourced LRMs, including DeepSeek-R1-1.5B and DeepSeek-R1-7B \citep{DeepSeek-R1}. All models are trained on the DeepScaleR dataset \citep{deepscaler2025}.

\paragraph{Implementation Details}
During rollout, we use a sampling temperature of 1.0 and a top-\(p\) value of 0.95, with the maximum output length set to 3{,}500 tokens. 
Training details are provided in the Appendix~\ref{app:Experimental Setup Details}.

\paragraph{Evaluation}
We evaluate on five widely used mathematical reasoning benchmarks: GSM8K \citep{gsm8k}, MATH-500 \citep{math500}, AMC23 \citep{AMC2023}, AIME24 \citep{AIME2024}, and AIME25 \citep{AIME2025}. We perform multi-sample evaluation by sampling \(N\) outputs per question, with \(N=64\) for AMC23, AIME24, and AIME25, and \(N=16\) for the remaining datasets.

\paragraph{Metrics}
We report five evaluation metrics: accuracy (\textbf{Acc}, in \%), \textbf{Pass@\(N\)} (in \%), average response length in tokens (\textbf{Tok}), compression rate (\textbf{CR}, in \%), and token efficiency (\textbf{Eff}). 
% \textbf{Pass@\(N\)} measures the fraction of problems for which at least one correct solution is obtained among \(N\) independently sampled outputs.
\textbf{Eff} represents the ratio of accuracy to average response length (\(\mathrm{Eff} = \mathrm{Acc} / \mathrm{Tok}\), in \%), and reflects the trade-off between correctness and efficiency. Details of the remaining metrics are provided in Appendix~\ref{app:Experimental Setup Details}.

% \begin{figure}[t]
%   \centering
%   \includegraphics[width=\linewidth]{Plot/temp_ablation_math500.pdf}
%   \caption{Training dynamics of the 1.5B model on MATH-500 under different rollout temperatures (0.3, 0.6, and 1.0).}
%   \label{fig:varying_temperature}
% \end{figure}

% \begin{figure}[t]
%   \centering
%   \includegraphics[width=\linewidth]{Plot/rolloutn_ablation_math500.pdf}
%   \caption{Training dynamics of the 1.5B model on MATH-500 under different rollout numbers (temperature = 1.0).}
%   \label{fig:varying_rolloutn}
% \end{figure}

% \begin{figure}[t]
%   \centering
%   \includegraphics[width=\linewidth]{Plot/loss_agg_ablation.pdf}
%   \caption{Training dynamics with and without length bias correction for the 1.5B model on MATH-500.}
%   \label{fig:correction_bias}
% \end{figure}

% \begin{figure}[t]
%   \centering
%   \includegraphics[width=\linewidth]{Plot/max_length_ablation_1.5b.pdf}
%   \caption{Training dynamics with and without length bias correction for the 1.5B model on MATH-500.}
%   \label{fig:max_len_ablation}
% \end{figure}

\paragraph{Baselines}
% We compare on-policy SFT with representative state-of-the-art methods for efficient reasoning.
The baselines fall into three major categories:
\begin{itemize}[leftmargin=*]
    \item \textbf{Training-free methods}  
    % These approaches improve reasoning efficiency at inference time without additional training.
    \emph{Chain of Draft (CoD)} \citep{CoD} is a prompting strategy that encourages models to generate concisely at each step; the prompt template is detailed in Appendix~\ref{app:prompt_cod}.
    \emph{Dynamic Early Exit in Reasoning (DEER)} \citep{deer} terminates the reasoning process when the model exhibits high confidence.

   \item \textbf{SFT-based methods}  
    % These methods construct compressed training data and fine-tune the model accordingly.
    \emph{Concise Reasoning Self-Training (CRST)} \citep{crst} fine-tunes models on concise reasoning paths obtained via best-of-\(N\) sampling.
    \emph{TokenSkip} \citep{tokenskip} prunes unimportant tokens to construct compressed training data; see Appendix~\ref{TokenSkip Implementation Details} for implementation details.
    \emph{StepEntropy} \citep{step_entropy} replaces the lowest-entropy \(80\%\) of reasoning steps with a special \texttt{[skip]} token.

    % \item \textbf{RL-based methods}
    % \emph{ThinkPrune} \citep{thinkprune} assigns a reward of $1$ to responses that are both correct and within a predefined maximum length, and $0$ otherwise; We set the maximum length to 3{,}500 for a fair comparison with our method.
    % \emph{O1-Pruner} \citep{O1-PRUNER} introduces a Length-Harmonizing Reward computed as the ratio between the CoT length of a reference model and that of the current policy.
    % \emph{L1} \citep{L1} trains models to satisfy user-specified length constraints; we report the \emph{L1-max} variant, which requires the output to be no longer than the specified target length, with the target length set to 3{,}500.
    % \emph{ER-RL} \citep{er_rl} assigns rewards only to correct responses, with shorter outputs receiving higher rewards.
    % \emph{LASER} \citep{laser} employs a step-function reward based on a target length, regardless of response correctness, and dynamically adjusts the target length during training.
    \item \textbf{RL-based methods}  
    \emph{ThinkPrune} \citep{thinkprune} assigns a reward of \(1\) to responses that are correct within a predefined length limit and \(0\) otherwise.
    \emph{O1-Pruner} \citep{O1-PRUNER} introduces a length-harmonizing reward computed as the ratio between the CoT length of a reference model and that of the current policy.
    \emph{L1} \citep{L1} trains models to satisfy user-specified length constraints; implementation details are provided in the Appendix~\ref{TokenSkip Implementation Details}.
    \emph{ER-RL} \citep{er_rl} assigns rewards only to correct responses, with shorter outputs receiving higher rewards.
    \emph{LASER} \citep{laser} employs a step-function reward based on a target length. 
    % and dynamically adjusts the target length during training.
\end{itemize}

% \begin{figure*}[t]
%   \centering
%   \begin{minipage}{0.32\textwidth}
%     \centering
%     \includegraphics[width=\linewidth]{Plot/temp_ablation_math500.pdf}
%     \caption{Training dynamics of the 1.5B model on MATH-500 under different rollout temperatures (0.3, 0.6, and 1.0).}
%     \label{varying_temperature}
%   \end{minipage}\hfill
%   \begin{minipage}{0.32\textwidth}
%     \centering
%     \includegraphics[width=\linewidth]{Plot/rolloutn_ablation_math500.pdf}
%     \caption{Training dynamics of the 1.5B model on MATH-500 under different rollout numbers (temperature = 1.0).}
%     \label{varying_rolloutn}
%   \end{minipage}\hfill
%   \begin{minipage}{0.32\textwidth}
%     \centering
%     \includegraphics[width=\linewidth]{Plot/loss_agg_ablation.pdf}
%     \caption{Training dynamics with and without length bias correction for the 1.5B model on MATH-500.}
%     \label{correction_bias}
%   \end{minipage}
%   \label{fig:three_in_one}
% \end{figure*}

\section{Experimental Results}
\label{Experimental Results}

\subsection{Main Experiment}
\paragraph{RL $>$ training-free $>$ SFT for efficient reasoning}
We summarize several key observations from our experiments.
(i) As shown in \autoref{main_table}, across both the 1.5B and 7B scales, RL-based methods consistently achieve superior accuracy–efficiency trade-offs, as reflected by higher Eff scores.
(ii) Somewhat surprisingly, training-free methods consistently outperform SFT-based approaches in terms of efficiency trade-offs, indicating that not all training paradigms equally improve efficient reasoning.
(iii) Although training-free methods preserve accuracy well, they struggle to achieve substantial compression on more challenging datasets. For instance, on the 7B model, CoD maintains accuracy on AIME25 but reduces generation length by only 2\%, whereas on simpler datasets such as GSM8K it achieves up to a 35.5\% reduction.
(iv) We observe that methods such as TokenSkip and StepEntropy, which prune tokens or steps from self-generated CoT traces during training, suffer from severe performance degradation. In contrast, CRST, which performs SFT using complete, fluent, and concise reasoning traces, achieves markedly better results. This suggests that training on coherent and well-formed reasoning sentences is essential for effective efficient reasoning.
\begin{figure*}[t]
  \centering
  \begin{minipage}[t]{0.24\linewidth}
    \centering
    \includegraphics[width=\linewidth]{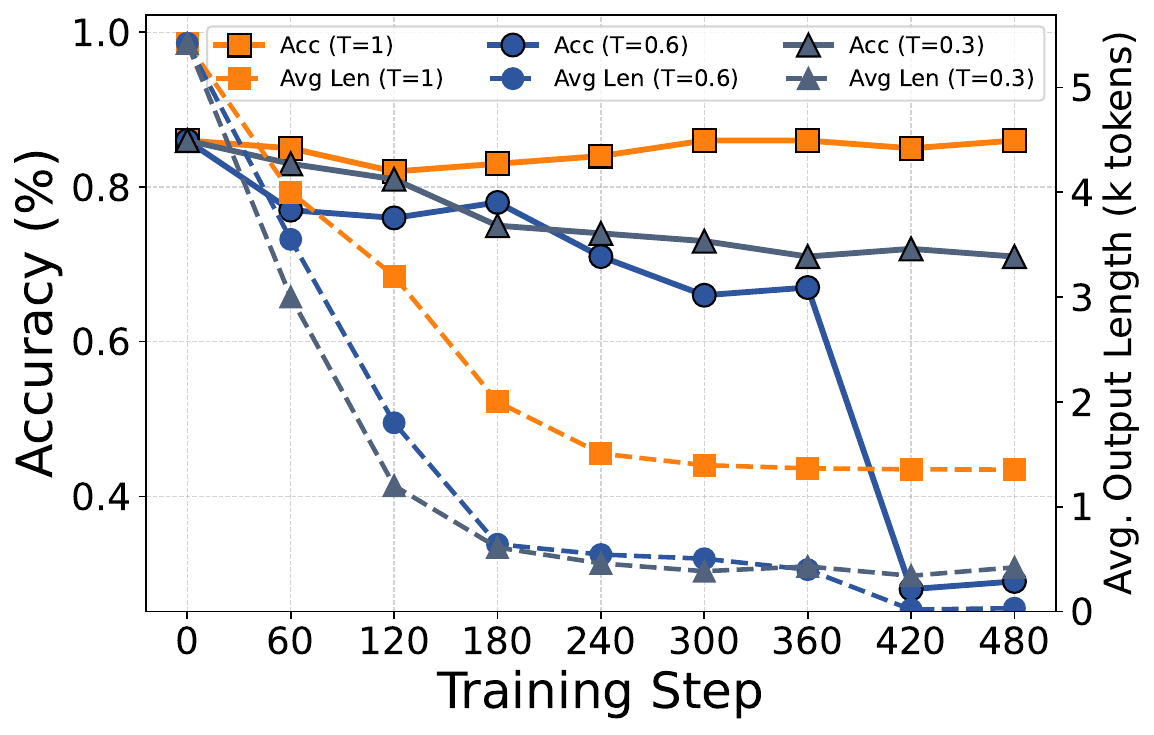}
    \vspace{-19pt}
    \captionof{figure}{Training dynamics of the 1.5B model on MATH-500 under different rollout temperatures (0.3, 0.6, and 1.0).}
    \label{fig:varying_temperature}
  \end{minipage}\hfill
  \begin{minipage}[t]{0.24\linewidth}
    \centering
    \includegraphics[width=\linewidth]{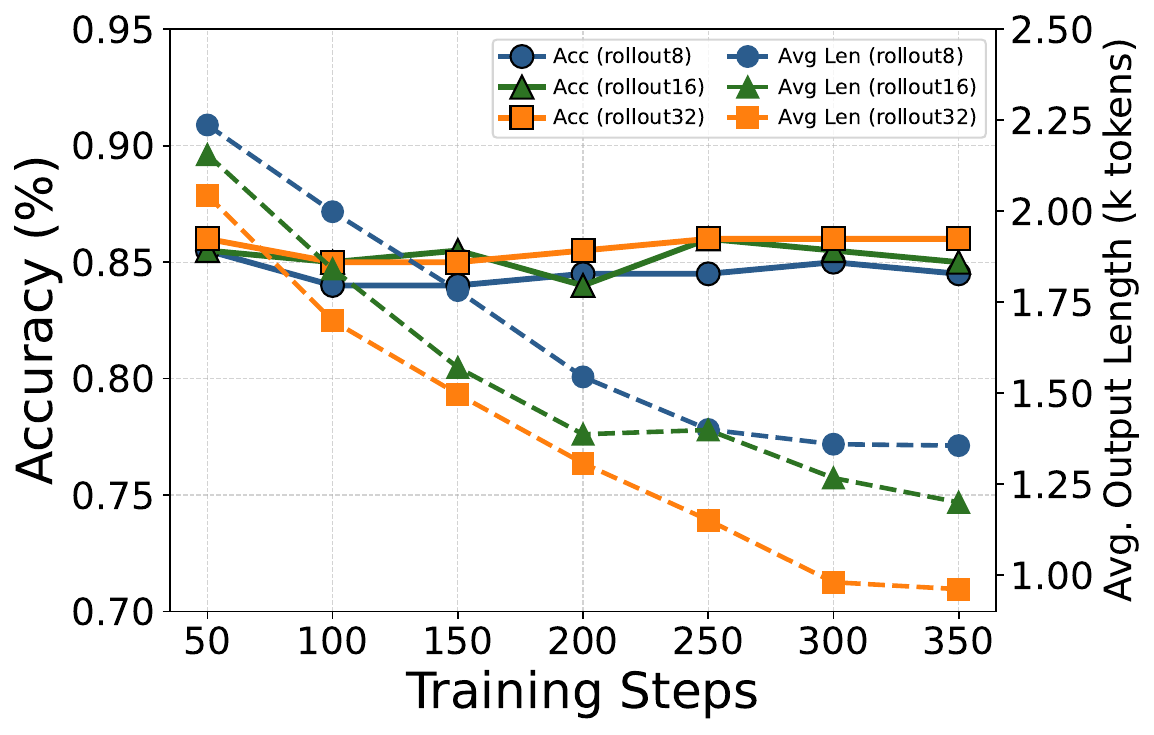}
    \vspace{-19pt}
    \captionof{figure}{Training dynamics of the 1.5B model on MATH-500 under different rollout numbers (temperature = 1.0).}
    \label{fig:varying_rolloutn}
  \end{minipage}\hfill
  \begin{minipage}[t]{0.24\linewidth}
    \centering
    \includegraphics[width=\linewidth]{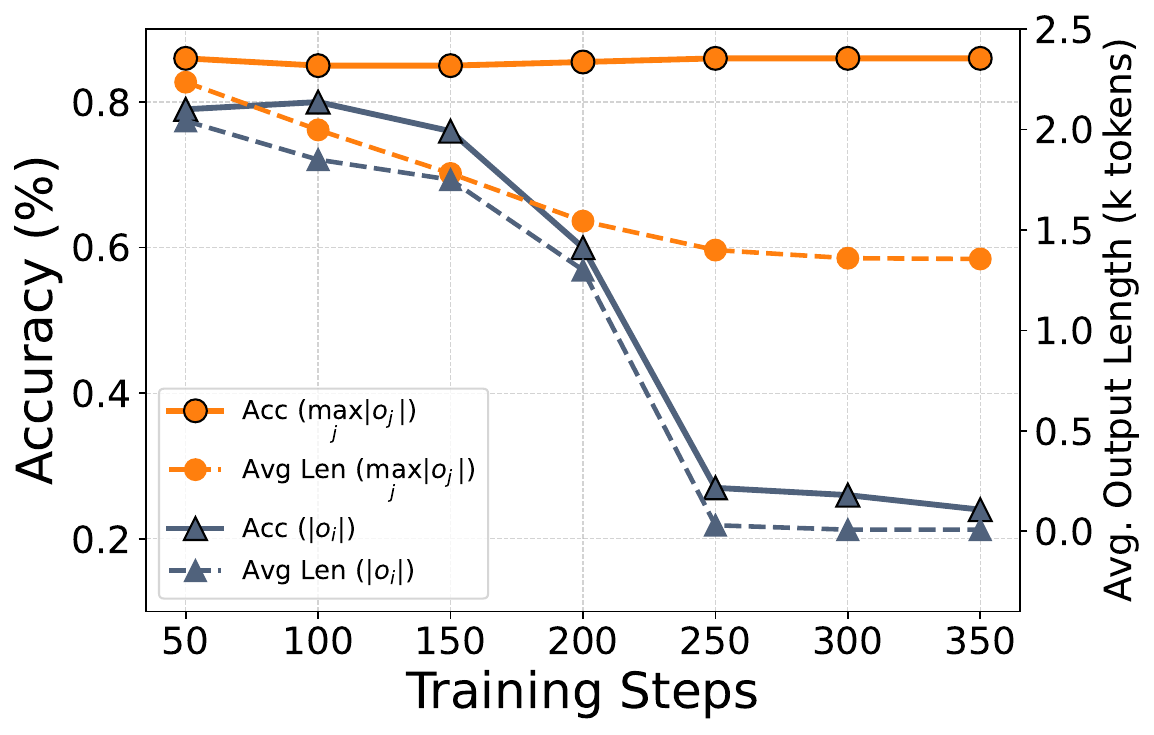}
    \vspace{-19pt}
    \captionof{figure}{Training dynamics with and without length bias correction for the 1.5B model on MATH-500.}
    \label{fig:correction_bias}
  \end{minipage}\hfill
  \begin{minipage}[t]{0.24\linewidth}
    \centering
    \includegraphics[width=\linewidth]{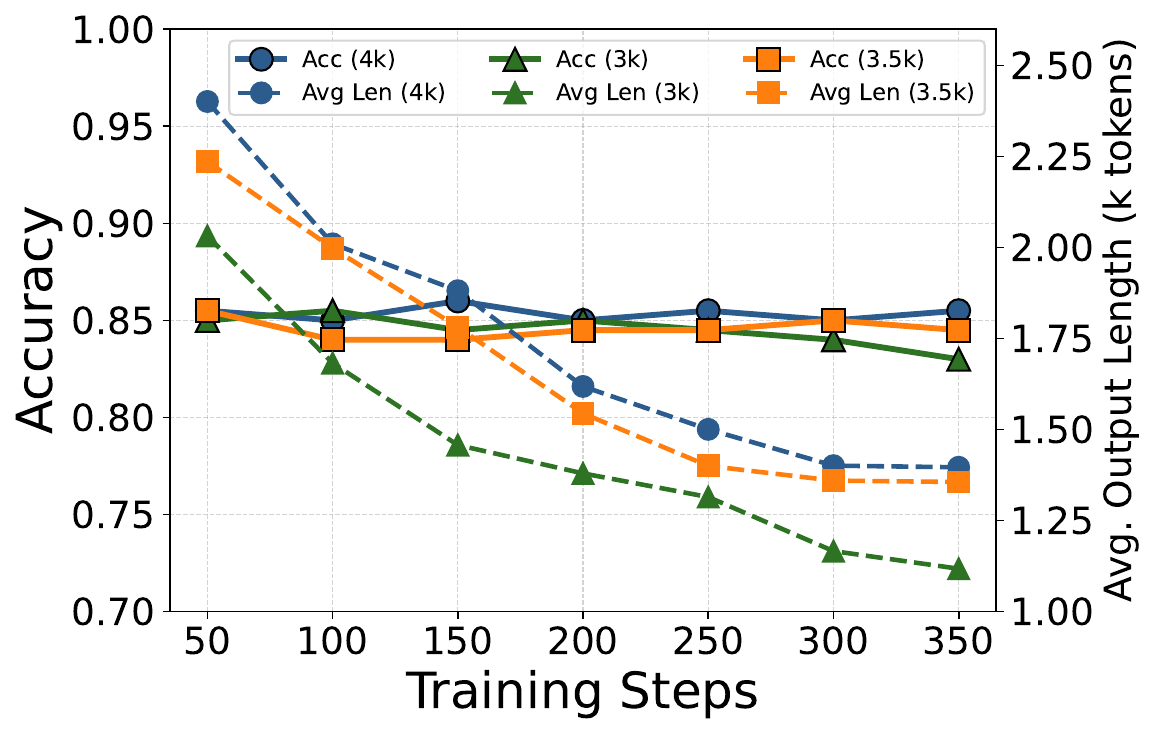}
    \vspace{-19pt}
    \captionof{figure}{Training dynamics with and without length bias correction for the 1.5B model on MATH-500.}
    \label{fig:max_len_ablation}
  \end{minipage}
\end{figure*}
\paragraph{On-policy SFT exhibits surprisingly strong effectiveness}
Despite its simplicity and the fact that on-policy SFT does not rely on reward signals, it \emph{achieves a superior performance--efficiency trade-off compared to existing state-of-the-art RL-based efficient reasoning methods} at both the 1.5B and 7B scales. As shown in \autoref{main_table}, on the 1.5B model, on-policy SFT attains an accuracy of 59.9\%, outperforming all baselines and the original CoT approach, while generating only 2{,}186 tokens on average—an approximately \textbf{80}\% reduction compared to the original model’s 10{,}178 tokens. When evaluated using Eff, which captures the trade-off between correctness and reasoning efficiency, our method achieves an Eff score of 2.7\%, exceeding the strongest RL-based baseline, L1 (2.55\%). A similar trend holds for the 7B model, where on-policy SFT attains the highest Eff score of 2.97\% and achieves an approximately \textbf{70}\% reduction in generation length, again surpassing all RL-based methods. In addition, \autoref{fig:acc_vs_tokens} compares the performance of on-policy SFT with all other baselines on AIME24 and MATH-500 under varying generation length budgets, where we observe that our method consistently lies on the Pareto frontier.

We further report\textit{ two intriguing findings}: 
(i) Despite using a truncation reward that assigns zero reward to responses exceeding 3{,}500 tokens, our method remains effective even when the average generation length surpasses this threshold. For example, on the 7B model evaluated on AIME25, on-policy SFT achieves 40.0\% accuracy and outperforms all baselines and the original CoT at an average length of 4{,}039 tokens. This behavior suggests strong generalization, as reasoning patterns learned from correct samples within the length-constrained regime transfer to longer responses.
(ii) Recent work emphasizes the role of negative samples in preventing entropy collapse and improving exploration, often yielding higher Pass@\(N\) scores \citep{retaining_by_doing, sample_polarity}. In contrast, our method optimizes exclusively on positive samples yet still achieves competitive Pass@\(N\) performance. On the 1.5B model, on-policy SFT attains a Pass@\(N\) of 73.6\%, slightly exceeding the original CoT (73.5\%) and trailing only LASER (74.8\%); on the 7B model, it reaches 85.7\%, outperforming the original CoT (84.6\%) and narrowly underperforming o1-pruner (86.1\%). We conjecture that negative samples may play a different role in multi-reward optimization than in settings focused solely on enhancing reasoning ability.

\paragraph{On-Policy SFT improves training efficiency}
Having established the inference-time efficiency of on-policy SFT, we next examine its training efficiency. 
% Owing to its simplified optimization objective, on-policy SFT avoids unnecessary implementation overhead and yields substantial gains in training efficiency. 
Specifically, we compare our method with the most competitive RL-based baseline ThinkPrune, which also relies on a truncation reward, in terms of average GPU memory consumption and wall-clock time per training step (including rollout generation and parameter updates) under varying rollout numbers, as shown in \autoref{fig:gpu_memory}. All experiments are conducted on a single 94GB H100 GPU without gradient accumulation. 
Across all rollout settings, on-policy SFT consistently reduces both time and memory costs by approximately \textbf{50}\% per training step relative to the RL-based method. 
The improved training efficiency of on-policy SFT stems from two factors.
First, it avoids maintaining a reference model for KL-divergence computation, reducing GPU memory usage and computational overhead. 
Second, training is performed only on filtered on-policy samples. 
We argue that this filtering is critical to the superior performance of on-policy SFT over ThinkPrune, since truncation-based rewards combined with group-wise normalization can penalize responses that would otherwise be correct when their length slightly exceeds the maximum constraint. 
Notably, beyond per-step efficiency, our method also accelerates convergence. Under identical training configurations, on-policy SFT converges in substantially fewer optimization steps. As shown in \autoref{fig:on_policy_sft_vs_thinkprune}, the average output length stabilizes after approximately 300 steps, whereas ThinkPrune requires around 500 steps to reach comparable stability—corresponding to a \textbf{40}\% reduction in training steps.
Consequently, the overall training process incurs significantly lower memory usage and wall-clock time. In practice, our method requires only about \textbf{30}\% of the wall-clock time needed by the RL-based baseline to converge.

\paragraph{On-policy SFT exhibits superior length control}
Beyond favorable efficiency–performance trade-offs, a well-behaved reasoning model should also exhibit strong length control, yielding stable response lengths across repeated samples.
To quantify this property, we measure the coefficient of variation of generation lengths, defined as the standard deviation normalized by the mean length \citep{coefficient_of_variation}; its formal definition is provided in Appendix~\ref{coefficient of variation}. Intuitively, smaller values indicate more consistent generation lengths and, consequently, stronger length control.
For the 1.5B model, we report results averaged over AIME24 and compare on-policy SFT with the RL-based baseline ThinkPrune, as shown in \autoref{fig:std_plot}. On-policy SFT produces shorter average responses and consistently exhibits stronger length control under varying token budgets.
% Notably, as the token budget increases, the gap in length variation between on-policy SFT and ThinkPrune further widens. Moreover, on-policy SFT exhibits a decreasing coefficient of variation as the generation length increases, indicating more stable length behavior under larger budgets. Together, these results indicate \textit{substantially stronger length control for on-policy SFT than for RL-based methods}.
We attribute this advantage to removing group-wise reward normalization, which avoids penalizing potentially correct answers and helps preserve the model’s confidence during generation.

These results suggest that the field may have introduced some unnecessary complexity, and that simpler alternatives can already achieve strong accuracy--efficiency Pareto frontiers with substantially fewer training resources. 
% Accordingly, we advocate a methodological shift toward simplicity: \textit{carefully analyzing the target setting, starting from an appropriate simple baseline, and introducing additional complexity only when such a baseline demonstrably fails}.

\subsection{The Key Driver of Effectiveness: On-Policy Data}
\label{The Key Driver of Effectiveness: On-Policy Data}

% A natural question is why SFT-based baselines fall short of on-policy SFT, even though both adopt the same SFT loss objective without reward dependence. For example, CRST fine-tunes on a fixed set of self-generated shortest correct trajectories, with a dataset size matched to that used by on-policy SFT. While this approach preserves performance, it achieves only limited compression. We identify the key distinction as lying in the data generation paradigm. In CRST, training data are generated once and then kept fixed throughout optimization, resulting in an off-policy setting. In contrast, on-policy SFT continually updates its training data by sampling from the current policy, thereby operating in an on-policy manner. This observation motivates a closer examination of the role of on-policy data in driving the effectiveness of our approach.
A natural question is why SFT-based baselines underperform on-policy SFT, despite sharing the same reward-free SFT objective. 
% For instance, CRST fine-tunes on a fixed set of self-generated shortest correct trajectories, with a dataset size matched to on-policy SFT.
% While this preserves accuracy, it yields only limited compression.
We identify the key distinction as lying in the data generation paradigm: SFT-based methods operate in an off-policy setting with a fixed dataset, whereas on-policy SFT continually resamples training data from the current policy.
This distinction motivates a closer examination of the role of on-policy data in our approach.

% To this end, we construct an off-policy variant for comparison. In this variant, we iteratively generate training datasets, where in each iteration the number of questions matches that used by on-policy SFT over 50 training steps, using the same rollout and filtering procedure: for each question, we sample 8 responses and retain only those that are correct and within the maximum length of 3{,}500 tokens. The resulting dataset is used to train the model for one epoch, and this data generation and training procedure is repeated for seven iterations, corresponding to the total number of training samples used over 350 on-policy training steps. We report evaluation results on MATH-500 for the 7B model in \autoref{fig:on_policy_and_off_policy}. As shown, on-policy SFT rapidly reduces the average generation length and converges to a stable regime while maintaining accuracy, whereas the off-policy variant exhibits only a gradual decrease in length accompanied by a sharp degradation in performance.
% In light of these results, we argue that \textit{the primary driver of effectiveness in both RL-based efficient reasoning and on-policy SFT is the use of on-policy data}, rather than complex optimization objectives or reward shaping strategies, which may even degrade performance. This observation is consistent with recent findings showing that the effectiveness of RL is largely attributable to on-policy data rather than sophisticated algorithmic choices~\citep{retaining_by_doing,rl_razor}.
To this end, we construct an off-policy variant for comparison. In each iteration, we generate a training dataset using the same rollout and filtering procedure as on-policy SFT—sampling 8 responses per question and retaining only correct responses within the 3{,}500-token limit—while matching the total number of questions used over 50 on-policy training steps. The model is then trained on this fixed dataset for one epoch, and the procedure is repeated for seven iterations, corresponding to 350 on-policy training steps in total.
We report results on MATH-500 for the 7B model in \autoref{fig:on_policy_and_off_policy}. As shown, on-policy SFT rapidly reduces generation length and converges to a stable regime while preserving accuracy, whereas the off-policy variant achieves only modest length reduction accompanied by substantial performance degradation.
These results suggest that the primary driver of effectiveness in both RL-based efficient reasoning and on-policy SFT is the use of on-policy data, rather than complex optimization objectives or reward shaping, which may even be detrimental. This conclusion aligns with recent findings that attribute the effectiveness of RL to on-policy data rather than sophisticated algorithmic choices \citep{retaining_by_doing,self_distillation_enables_continual_learning,self_distilled_reasoner,rl_via_self_distillation}.

\subsection{Practical Training Guidelines}
\label{Practical Training Guidelines}
To promote stable training, we provide practical training guidelines for on-policy SFT, together with principled explanations of the underlying rationale.
% Unless otherwise specified, all analyses in this subsection use the 1.5B model and are evaluated on MATH-500.

\paragraph{Rollout temperature should be set to 1.0}
We illustrate the training dynamics under different rollout temperatures in \autoref{fig:varying_temperature}, considering values of 0.3, 0.6, and 1.0.
We observe that, except for temperature 1.0, lower temperatures consistently lead to excessive length reduction, accompanied by performance degradation or even collapse. We further evaluate higher temperatures of 1.2 and 1.5, as shown in \autoref{fig:training_loss_temp}, under which training becomes unstable, with the loss monotonically increasing and failing to converge. This behavior arises because temperatures other than 1.0 violate the on-policy assumption. We provide a theoretical justification for this claim, with a detailed proof presented in the Appendix~\ref{On-Policy Requirement for Rollout Temperature}.
% \paragraph{Rollout temperature should be set to 1.0}
% We analyze training dynamics under different rollout temperatures in \autoref{fig:varying_temperature}, considering 0.3, 0.6, and 1.0. Except for 1.0, lower temperatures cause excessive length reduction accompanied by performance degradation or collapse. Conversely, higher temperatures (1.2 and 1.5) lead to unstable training, with monotonically increasing loss and failure to converge, as shown in \autoref{fig:training_loss_temp}. This instability arises because temperatures other than 1.0 violate the on-policy assumption. A theoretical justification is provided in Appendix~\ref{On-Policy Requirement for Rollout Temperature}.

% \paragraph{On-policy SFT benefits from scaling the rollout number}
% Under a rollout temperature of 1.0, we examine the training dynamics with varying rollout numbers. As shown in \autoref{fig:varying_rolloutn}, all rollout configurations exhibit stable training behavior. Moreover, increasing the rollout number consistently leads to shorter average response lengths while maintaining accuracy. These results indicate that, provided the rollout temperature is set to 1.0 and sufficient computational resources are available, scaling the rollout number is an effective strategy for improving the performance of on-policy SFT.
\paragraph{On-policy SFT benefits from scaling the rollout number}
With the rollout temperature fixed at 1.0, we examine training dynamics under varying rollout numbers. As shown in \autoref{fig:varying_rolloutn}, all configurations remain stable. Moreover, increasing the rollout number consistently yields shorter responses while preserving accuracy. These results suggest that, given a rollout temperature of 1.0 and sufficient computational resources, scaling the rollout number is an effective way to improve on-policy SFT performance.

% \paragraph{Length bias correction is necessary}
% We study the bias-correction scheme described in \autoref{On-Policy SFT subsection}, where $|o_i|$ is replaced by the maximum response length within a batch, $\max_j |o_j|$. We visualize the training dynamics under both settings in \autoref{fig:correction_bias}. As shown, directly using $|o_i|$ leads to training collapse: both generation length and accuracy rapidly decay and approach zero after approximately 250 training steps. In contrast, adopting $\max_j |o_j|$ yields stable training dynamics. This behavior arises because using $|o_i|$ assigns larger gradient updates to shorter responses, making training overly sensitive to noise and prone to converging to suboptimal solutions or even collapsing.
\paragraph{Length bias correction is necessary}
We analyze the bias-correction scheme in \autoref{On-Policy SFT subsection}, which replaces \(|o_i|\) with the batch maximum \(\max_j |o_j|\). Training dynamics under both settings are shown in \autoref{fig:correction_bias}. Using \(|o_i|\) leads to training collapse, with both generation length and accuracy rapidly decaying to near zero after roughly 250 steps. In contrast, adopting \(\max_j |o_j|\) yields stable training. 
% This occurs because \(|o_i|\) assigns larger gradients to shorter responses, making training overly sensitive to noise and prone to suboptimal convergence or collapse.

% \paragraph{Maximum length involves a trade-off between performance and training cost}
% We analyze the training dynamics under different maximum generation lengths, as shown in \autoref{fig:max_len_ablation}. A maximum length of 3k tokens results in lower training cost and the fastest reduction in average response length, but with a slight performance degradation. In contrast, a maximum length of 4k tokens incurs the highest training cost and leads to a slower yet more stable reduction in response length, while achieving the best overall performance. Our chosen setting of 3{,}500 tokens strikes a balance, achieving performance comparable to the 4k setting with moderate training cost.
% \paragraph{Length limit trades off performance and training cost}
% We examine training dynamics under different maximum generation lengths in \autoref{fig:max_len_ablation}. A maximum length of 3k tokens yields the lowest training cost and the fastest reduction in response length, but with slight performance degradation. In contrast, 4k tokens incur the highest cost and produce slower but more stable length reduction, while achieving the best performance. Our chosen setting of 3{,}500 tokens balances these trade-offs, matching the performance of 4k with moderate training cost.
\paragraph{Length limit trades off performance and training cost}
We analyze training dynamics under different maximum generation lengths in \autoref{fig:max_len_ablation}. A limit of 3k tokens yields the lowest training cost and fastest length reduction, but with slight performance degradation, whereas 4k tokens incur the highest cost and achieve the best performance with slower, more stable length reduction. Our chosen setting of 3.5k tokens strikes a balance, matching 4k performance at moderate training cost.

\section{Conclusion}
We revisit RL-based efficient reasoning from a minimalist perspective and show that much of the complexity inherited from GRPO, including KL regularization, group-wise normalization, and sophisticated reward shaping, may not be necessary in this setting. By deliberately removing these components, we derive a simple, reward-free training procedure that is mathematically equivalent to SFT on on-policy data. Despite its simplicity, on-policy SFT achieves strong accuracy–efficiency trade-offs, improves training efficiency, and exhibits robust length control across benchmarks and model scales. Overall, our findings suggest that future progress in efficient reasoning may benefit from prioritizing simplicity and principled design over increasing complexity.

% \input{Input/Format}

% Authors are kindly asked to make their submissions as accessible as possible
% for everyone including people with disabilities and sensory or neurological
% differences. Tips of how to achieve this and what to pay attention to will be
% provided on the conference website \url{http://icml.cc/}.

% \section*{Software and Data}

% If a paper is accepted, we strongly encourage the publication of software and
% data with the camera-ready version of the paper whenever appropriate. This can
% be done by including a URL in the camera-ready copy. However, \textbf{do not}
% include URLs that reveal your institution or identity in your submission for
% review. Instead, provide an anonymous URL or upload the material as
% ``Supplementary Material'' into the OpenReview reviewing system. Note that
% reviewers are not required to look at this material when writing their review.

% % Acknowledgements should only appear in the accepted version.
% \section*{Acknowledgements}

% \textbf{Do not} include acknowledgements in the initial version of the paper

% If a paper is accepted, the final camera-ready version can (and usually should)
% include acknowledgements.  Such acknowledgements should be placed at the end of
% the section, in an unnumbered section that does not count towards the paper
% page limit. Typically, this will include thanks to reviewers who gave useful
% comments, to colleagues who contributed to the ideas, and to funding agencies
% and corporate sponsors that provided financial support.

\section*{Impact Statement}

% Authors are \textbf{required} to include a statement of the potential broader
% impact of their work, including its ethical aspects and future societal
% consequences. This statement should be in an unnumbered section at the end of
% the paper (co-located with Acknowledgements -- the two may appear in either
% order, but both must be before References), and does not count toward the paper
% page limit. In many cases, where the ethical impacts and expected societal
% implications are those that are well established when advancing the field of
% Machine Learning, substantial discussion is not required, and a simple
% statement such as the following will suffice:

% ``This paper presents work whose goal is to advance the field of Machine
% Learning. There are many potential societal consequences of our work, none
% which we feel must be specifically highlighted here.''

% The above statement can be used verbatim in such cases, but we encourage
% authors to think about whether there is content which does warrant further
% discussion, as this statement will be apparent if the paper is later flagged
% for ethics review.

This work contributes to more efficient training and deployment of LRMs by showing that competitive accuracy–efficiency trade-offs can be achieved through simplified and principled training objectives. By reducing computational and memory requirements, the proposed approach may lower the resource barrier for developing and applying reasoning-capable language models, potentially enabling broader access in research and educational settings. At the same time, the method does not introduce new capabilities beyond improved efficiency, and thus does not pose additional risks compared to existing large language models. As with prior work on reasoning models, responsible use and deployment remain important considerations.

% % In the unusual situation where you want a paper to appear in the
% % references without citing it in the main text, use \nocite

\bibliography{main}

@misc{LRM_survey,
      title={Towards Large Reasoning Models: A Survey of Reinforced Reasoning with Large Language Models}, 
      author={Fengli Xu and Qianyue Hao and Zefang Zong and Jingwei Wang and Yunke Zhang and Jingyi Wang and Xiaochong Lan and Jiahui Gong and Tianjian Ouyang and Fanjin Meng and Chenyang Shao and Yuwei Yan and Qinglong Yang and Yiwen Song and Sijian Ren and Xinyuan Hu and Yu Li and Jie Feng and Chen Gao and Yong Li},
      year={2025},
      eprint={2501.09686},
      archivePrefix={arXiv},
      primaryClass={cs.AI},
      url={https://arxiv.org/abs/2501.09686}, 
}

@misc{openai,
  author       = {{OpenAI}},
  title        = {Learning to Reason with LLMs},
  year         = {2024},
  howpublished = {\url{https://openai.com/index/learning-to-reason-with-llms/}},
  note         = {OpenAI}
}

@article{DeepSeek-R1,
   title={DeepSeek-R1 incentivizes reasoning in LLMs through reinforcement learning},
   volume={645},
   ISSN={1476-4687},
   url={http://dx.doi.org/10.1038/s41586-025-09422-z},
   DOI={10.1038/s41586-025-09422-z},
   number={8081},
   journal={Nature},
   publisher={Springer Science and Business Media LLC},
   author={Guo, Daya and Yang, Dejian and Zhang, Haowei and Song, Junxiao and Wang, Peiyi and Zhu, Qihao and Xu, Runxin and Zhang, Ruoyu and Ma, Shirong and Bi, Xiao and Zhang, Xiaokang and Yu, Xingkai and Wu, Yu and Wu, Z. F. and Gou, Zhibin and Shao, Zhihong and Li, Zhuoshu and Gao, Ziyi and Liu, Aixin and Xue, Bing and Wang, Bingxuan and Wu, Bochao and Feng, Bei and Lu, Chengda and Zhao, Chenggang and Deng, Chengqi and Ruan, Chong and Dai, Damai and Chen, Deli and Ji, Dongjie and Li, Erhang and Lin, Fangyun and Dai, Fucong and Luo, Fuli and Hao, Guangbo and Chen, Guanting and Li, Guowei and Zhang, H. and Xu, Hanwei and Ding, Honghui and Gao, Huazuo and Qu, Hui and Li, Hui and Guo, Jianzhong and Li, Jiashi and Chen, Jingchang and Yuan, Jingyang and Tu, Jinhao and Qiu, Junjie and Li, Junlong and Cai, J. L. and Ni, Jiaqi and Liang, Jian and Chen, Jin and Dong, Kai and Hu, Kai and You, Kaichao and Gao, Kaige and Guan, Kang and Huang, Kexin and Yu, Kuai and Wang, Lean and Zhang, Lecong and Zhao, Liang and Wang, Litong and Zhang, Liyue and Xu, Lei and Xia, Leyi and Zhang, Mingchuan and Zhang, Minghua and Tang, Minghui and Zhou, Mingxu and Li, Meng and Wang, Miaojun and Li, Mingming and Tian, Ning and Huang, Panpan and Zhang, Peng and Wang, Qiancheng and Chen, Qinyu and Du, Qiushi and Ge, Ruiqi and Zhang, Ruisong and Pan, Ruizhe and Wang, Runji and Chen, R. J. and Jin, R. L. and Chen, Ruyi and Lu, Shanghao and Zhou, Shangyan and Chen, Shanhuang and Ye, Shengfeng and Wang, Shiyu and Yu, Shuiping and Zhou, Shunfeng and Pan, Shuting and Li, S. S. and Zhou, Shuang and Wu, Shaoqing and Yun, Tao and Pei, Tian and Sun, Tianyu and Wang, T. and Zeng, Wangding and Liu, Wen and Liang, Wenfeng and Gao, Wenjun and Yu, Wenqin and Zhang, Wentao and Xiao, W. L. and An, Wei and Liu, Xiaodong and Wang, Xiaohan and Chen, Xiaokang and Nie, Xiaotao and Cheng, Xin and Liu, Xin and Xie, Xin and Liu, Xingchao and Yang, Xinyu and Li, Xinyuan and Su, Xuecheng and Lin, Xuheng and Li, X. Q. and Jin, Xiangyue and Shen, Xiaojin and Chen, Xiaosha and Sun, Xiaowen and Wang, Xiaoxiang and Song, Xinnan and Zhou, Xinyi and Wang, Xianzu and Shan, Xinxia and Li, Y. K. and Wang, Y. Q. and Wei, Y. X. and Zhang, Yang and Xu, Yanhong and Li, Yao and Zhao, Yao and Sun, Yaofeng and Wang, Yaohui and Yu, Yi and Zhang, Yichao and Shi, Yifan and Xiong, Yiliang and He, Ying and Piao, Yishi and Wang, Yisong and Tan, Yixuan and Ma, Yiyang and Liu, Yiyuan and Guo, Yongqiang and Ou, Yuan and Wang, Yuduan and Gong, Yue and Zou, Yuheng and He, Yujia and Xiong, Yunfan and Luo, Yuxiang and You, Yuxiang and Liu, Yuxuan and Zhou, Yuyang and Zhu, Y. X. and Huang, Yanping and Li, Yaohui and Zheng, Yi and Zhu, Yuchen and Ma, Yunxian and Tang, Ying and Zha, Yukun and Yan, Yuting and Ren, Z. Z. and Ren, Zehui and Sha, Zhangli and Fu, Zhe and Xu, Zhean and Xie, Zhenda and Zhang, Zhengyan and Hao, Zhewen and Ma, Zhicheng and Yan, Zhigang and Wu, Zhiyu and Gu, Zihui and Zhu, Zijia and Liu, Zijun and Li, Zilin and Xie, Ziwei and Song, Ziyang and Pan, Zizheng and Huang, Zhen and Xu, Zhipeng and Zhang, Zhongyu and Zhang, Zhen},
   year={2025},
   month=sep, pages={633–638} }

@misc{qwen3_technical_report,
      title={Qwen3 Technical Report}, 
      author={An Yang and Anfeng Li and Baosong Yang and Beichen Zhang and Binyuan Hui and Bo Zheng and Bowen Yu and Chang Gao and Chengen Huang and Chenxu Lv and Chujie Zheng and Dayiheng Liu and Fan Zhou and Fei Huang and Feng Hu and Hao Ge and Haoran Wei and Huan Lin and Jialong Tang and Jian Yang and Jianhong Tu and Jianwei Zhang and Jianxin Yang and Jiaxi Yang and Jing Zhou and Jingren Zhou and Junyang Lin and Kai Dang and Keqin Bao and Kexin Yang and Le Yu and Lianghao Deng and Mei Li and Mingfeng Xue and Mingze Li and Pei Zhang and Peng Wang and Qin Zhu and Rui Men and Ruize Gao and Shixuan Liu and Shuang Luo and Tianhao Li and Tianyi Tang and Wenbiao Yin and Xingzhang Ren and Xinyu Wang and Xinyu Zhang and Xuancheng Ren and Yang Fan and Yang Su and Yichang Zhang and Yinger Zhang and Yu Wan and Yuqiong Liu and Zekun Wang and Zeyu Cui and Zhenru Zhang and Zhipeng Zhou and Zihan Qiu},
      year={2025},
      eprint={2505.09388},
      archivePrefix={arXiv},
      primaryClass={cs.CL},
      url={https://arxiv.org/abs/2505.09388}, 
}

@article{ding2026llms,
  title={From LLMs to LRMs: Rethinking Pruning for Reasoning-Centric Models},
  author={Ding, Longwei and Zhao, Anhao and Ye, Fanghua and Chen, Ziyang and Shen, Xiaoyu},
  journal={arXiv preprint arXiv:2601.18091},
  year={2026}
}

@inproceedings{chen2025unveiling,
  title={Unveiling the key factors for distilling chain-of-thought reasoning},
  author={Chen, Xinghao and Sun, Zhijing and Wenjin, Guo and Zhang, Miaoran and Chen, Yanjun and Sun, Yirong and Su, Hui and Pan, Yijie and Klakow, Dietrich and Li, Wenjie and others},
  booktitle={Findings of the Association for Computational Linguistics: ACL 2025},
  pages={15094--15119},
  year={2025}
}

@misc{cot,
      title={Chain-of-Thought Prompting Elicits Reasoning in Large Language Models}, 
      author={Jason Wei and Xuezhi Wang and Dale Schuurmans and Maarten Bosma and Brian Ichter and Fei Xia and Ed Chi and Quoc Le and Denny Zhou},
      year={2023},
      eprint={2201.11903},
      archivePrefix={arXiv},
      primaryClass={cs.CL},
      url={https://arxiv.org/abs/2201.11903}, 
}

@misc{dr.grpo,
      title={Understanding R1-Zero-Like Training: A Critical Perspective}, 
      author={Zichen Liu and Changyu Chen and Wenjun Li and Penghui Qi and Tianyu Pang and Chao Du and Wee Sun Lee and Min Lin},
      year={2025},
      eprint={2503.20783},
      archivePrefix={arXiv},
      primaryClass={cs.LG},
      url={https://arxiv.org/abs/2503.20783}, 
}

@article{deepseekmath,
  title={Deepseekmath: Pushing the limits of mathematical reasoning in open language models},
  author={Shao, Zhihong and Wang, Peiyi and Zhu, Qihao and Xu, Runxin and Song, Junxiao and Bi, Xiao and Zhang, Haowei and Zhang, Mingchuan and Li, YK and Wu, Yang and others},
  journal={arXiv preprint arXiv:2402.03300},
  year={2024}
}

@misc{er_rl,
      title={Training Language Models to Reason Efficiently}, 
      author={Daman Arora and Andrea Zanette},
      year={2025},
      eprint={2502.04463},
      archivePrefix={arXiv},
      primaryClass={cs.LG},
      url={https://arxiv.org/abs/2502.04463}, 
}

@misc{thinkprune,
      title={ThinkPrune: Pruning Long Chain-of-Thought of LLMs via Reinforcement Learning}, 
      author={Bairu Hou and Yang Zhang and Jiabao Ji and Yujian Liu and Kaizhi Qian and Jacob Andreas and Shiyu Chang},
      year={2025},
      eprint={2504.01296},
      archivePrefix={arXiv},
      primaryClass={cs.CL},
      url={https://arxiv.org/abs/2504.01296}, 
}

@misc{overthinking,
      title={Do NOT Think That Much for 2+3=? On the Overthinking of o1-Like LLMs}, 
      author={Xingyu Chen and Jiahao Xu and Tian Liang and Zhiwei He and Jianhui Pang and Dian Yu and Linfeng Song and Qiuzhi Liu and Mengfei Zhou and Zhuosheng Zhang and Rui Wang and Zhaopeng Tu and Haitao Mi and Dong Yu},
      year={2025},
      eprint={2412.21187},
      archivePrefix={arXiv},
      primaryClass={cs.CL},
      url={https://arxiv.org/abs/2412.21187}, 
}

@article{tokenskip,
  title={Tokenskip: Controllable chain-of-thought compression in llms},
  author={Xia, Heming and Leong, Chak Tou and Wang, Wenjie and Li, Yongqi and Li, Wenjie},
  journal={arXiv preprint arXiv:2502.12067},
  year={2025}
}

@article{CoD,
  title={Chain of draft: Thinking faster by writing less},
  author={Xu, Silei and Xie, Wenhao and Zhao, Lingxiao and He, Pengcheng},
  journal={arXiv preprint arXiv:2502.18600},
  year={2025}
}

@misc{O1-PRUNER,
      title={O1-Pruner: Length-Harmonizing Fine-Tuning for O1-Like Reasoning Pruning}, 
      author={Haotian Luo and Li Shen and Haiying He and Yibo Wang and Shiwei Liu and Wei Li and Naiqiang Tan and Xiaochun Cao and Dacheng Tao},
      year={2025},
      eprint={2501.12570},
      archivePrefix={arXiv},
      primaryClass={cs.CL},
      url={https://arxiv.org/abs/2501.12570}, 
}

@article{KIMI,
  title={Kimi k1. 5: Scaling reinforcement learning with llms},
  author={Team, Kimi and Du, Angang and Gao, Bofei and Xing, Bowei and Jiang, Changjiu and Chen, Cheng and Li, Cheng and Xiao, Chenjun and Du, Chenzhuang and Liao, Chonghua and others},
  journal={arXiv preprint arXiv:2501.12599},
  year={2025}
}

@misc{aware_first,
      title={Aware First, Think Less: Dynamic Boundary Self-Awareness Drives Extreme Reasoning Efficiency in Large Language Models}, 
      author={Qiguang Chen and Dengyun Peng and Jinhao Liu and HuiKang Su and Jiannan Guan and Libo Qin and Wanxiang Che},
      year={2025},
      eprint={2508.11582},
      archivePrefix={arXiv},
      primaryClass={cs.CL},
      url={https://arxiv.org/abs/2508.11582}, 
}

@misc{correct_concise_and_complete,
      title={Correct, Concise and Complete: Multi-stage Training For Adaptive Reasoning}, 
      author={Nathanaël Carraz Rakotonirina and Ren Pang and Neha Anna John and Michael Bohlke-Schneider and Momchil Hardalov},
      year={2026},
      eprint={2601.02972},
      archivePrefix={arXiv},
      primaryClass={cs.CL},
      url={https://arxiv.org/abs/2601.02972}, 
}

@misc{dler_nvidia,
      title={DLER: Doing Length pEnalty Right - Incentivizing More Intelligence per Token via Reinforcement Learning}, 
      author={Shih-Yang Liu and Xin Dong and Ximing Lu and Shizhe Diao and Mingjie Liu and Min-Hung Chen and Hongxu Yin and Yu-Chiang Frank Wang and Kwang-Ting Cheng and Yejin Choi and Jan Kautz and Pavlo Molchanov},
      year={2025},
      eprint={2510.15110},
      archivePrefix={arXiv},
      primaryClass={cs.LG},
      url={https://arxiv.org/abs/2510.15110}, 
}

@misc{L1,
      title={L1: Controlling How Long A Reasoning Model Thinks With Reinforcement Learning}, 
      author={Pranjal Aggarwal and Sean Welleck},
      year={2025},
      eprint={2503.04697},
      archivePrefix={arXiv},
      primaryClass={cs.CL},
      url={https://arxiv.org/abs/2503.04697}, 
}

@misc{ppo,
      title={Proximal Policy Optimization Algorithms}, 
      author={John Schulman and Filip Wolski and Prafulla Dhariwal and Alec Radford and Oleg Klimov},
      year={2017},
      eprint={1707.06347},
      archivePrefix={arXiv},
      primaryClass={cs.LG},
      url={https://arxiv.org/abs/1707.06347}, 
}

@misc{retaining_by_doing,
      title={Retaining by Doing: The Role of On-Policy Data in Mitigating Forgetting}, 
      author={Howard Chen and Noam Razin and Karthik Narasimhan and Danqi Chen},
      year={2025},
      eprint={2510.18874},
      archivePrefix={arXiv},
      primaryClass={cs.LG},
      url={https://arxiv.org/abs/2510.18874}, 
}

@misc{token_budget_prompt,
      title={Token-Budget-Aware LLM Reasoning}, 
      author={Tingxu Han and Zhenting Wang and Chunrong Fang and Shiyu Zhao and Shiqing Ma and Zhenyu Chen},
      year={2025},
      eprint={2412.18547},
      archivePrefix={arXiv},
      primaryClass={cs.CL},
      url={https://arxiv.org/abs/2412.18547}, 
}

@misc{no_thinking,
      title={Reasoning Models Can Be Effective Without Thinking}, 
      author={Wenjie Ma and Jingxuan He and Charlie Snell and Tyler Griggs and Sewon Min and Matei Zaharia},
      year={2025},
      eprint={2504.09858},
      archivePrefix={arXiv},
      primaryClass={cs.AI},
      url={https://arxiv.org/abs/2504.09858}, 
}

@misc{step_entropy,
      title={Compressing Chain-of-Thought in LLMs via Step Entropy}, 
      author={Zeju Li and Jianyuan Zhong and Ziyang Zheng and Xiangyu Wen and Zhijian Xu and Yingying Cheng and Fan Zhang and Qiang Xu},
      year={2025},
      eprint={2508.03346},
      archivePrefix={arXiv},
      primaryClass={cs.AI},
      url={https://arxiv.org/abs/2508.03346}, 
}

@misc{laser,
      title={Learn to Reason Efficiently with Adaptive Length-based Reward Shaping}, 
      author={Wei Liu and Ruochen Zhou and Yiyun Deng and Yuzhen Huang and Junteng Liu and Yuntian Deng and Yizhe Zhang and Junxian He},
      year={2025},
      eprint={2505.15612},
      archivePrefix={arXiv},
      primaryClass={cs.CL},
      url={https://arxiv.org/abs/2505.15612}, 
}

@misc{xiaomi,
      title={Reinforcement Learning for Chain of Thought Compression with One-Domain-to-All Generalization}, 
      author={Hanyu Li and Jiangshan Duo and Bofei Gao and Hailin Zhang and Sujian Li and Xiaotie Deng and Liang Zhao},
      year={2025},
      eprint={2601.06052},
      archivePrefix={arXiv},
      primaryClass={cs.CL},
      url={https://arxiv.org/abs/2601.06052}, 
}

@article{REINFORCEMENT,
  title={Simple statistical gradient-following algorithms for connectionist reinforcement learning},
  author={Williams, Ronald J},
  journal={Machine learning},
  volume={8},
  number={3},
  pages={229--256},
  year={1992},
  publisher={Springer}
}

@misc{deepscaler2025,
  title  = {DeepScaleR: Surpassing O1-Preview with a 1.5B Model by Scaling RL},
  author = {Michael Luo and Sijun Tan and Justin Wong and Xiaoxiang Shi and William Y. Tang and Manan Roongta and Colin Cai and Jeffrey Luo and Li Erran Li and Raluca Ada Popa and Ion Stoica},
  note   = {Notion Blog},
  year   = {2025}
}

@inproceedings{omni,
  title={Omni-MATH: A Universal Olympiad Level Mathematic Benchmark for Large Language Models},
  author={Gao, Bofei and Song, Feifan and Yang, Zhe and Cai, Zefan and Miao, Yibo and Dong, Qingxiu and Li, Lei and Ma, Chenghao and Chen, Liang and Xu, Runxin and others},
  booktitle={The Thirteenth International Conference on Learning Representations},
  year={2025}
}

@misc{STILL,
      title={Imitate, Explore, and Self-Improve: A Reproduction Report on Slow-thinking Reasoning Systems}, 
      author={Yingqian Min and Zhipeng Chen and Jinhao Jiang and Jie Chen and Jia Deng and Yiwen Hu and Yiru Tang and Jiapeng Wang and Xiaoxue Cheng and Huatong Song and Wayne Xin Zhao and Zheng Liu and Zhongyuan Wang and Ji-Rong Wen},
      year={2024},
      eprint={2412.09413},
      archivePrefix={arXiv},
      primaryClass={cs.AI},
      url={https://arxiv.org/abs/2412.09413}, 
}

@article{gsm8k,
  title={Training verifiers to solve math word problems},
  author={Cobbe, Karl and Kosaraju, Vineet and Bavarian, Mohammad and Chen, Mark and Jun, Heewoo and Kaiser, Lukasz and Plappert, Matthias and Tworek, Jerry and Hilton, Jacob and Nakano, Reiichiro and others},
  journal={arXiv preprint arXiv:2110.14168},
  year={2021}
}

@inproceedings{math500,
  title={Let's verify step by step},
  author={Lightman, Hunter and Kosaraju, Vineet and Burda, Yuri and Edwards, Harrison and Baker, Bowen and Lee, Teddy and Leike, Jan and Schulman, John and Sutskever, Ilya and Cobbe, Karl},
  booktitle={The Twelfth International Conference on Learning Representations},
  year={2023}
}

@misc{AMC2023,
  author       = {{AMC2023}},
  title        = {American Mathematics Competitions},
  year         = {2023},
  url          = {https://artofproblemsolving.com/wiki/index.php/AMC_Problems_and_Solutions},
  urldate      = {2024-01-18}
}

@misc{AIME2024,
  author       = {{AIME2024}},
  title        = {American Invitational Mathematics Examination (AIME) 2024-I \& II},
  year         = {2024},
  howpublished = {\url{https://huggingface.co/datasets/Maxwell-Jia/AIME_2024}},
  note         = {Accessed: 2024-01-18}
}

@misc{AIME2025,
  author       = {{AIME2025}},
  title        = {American Invitational Mathematics Examination (AIME) 2025-I \& II},
  year         = {2025},
  howpublished = {\url{https://huggingface.co/datasets/opencompass/AIME2025}},
  note         = {Accessed: 2024-01-18}
}

@misc{deer,
      title={Dynamic Early Exit in Reasoning Models}, 
      author={Chenxu Yang and Qingyi Si and Yongjie Duan and Zheliang Zhu and Chenyu Zhu and Qiaowei Li and Minghui Chen and Zheng Lin and Weiping Wang},
      year={2025},
      eprint={2504.15895},
      archivePrefix={arXiv},
      primaryClass={cs.CL},
      url={https://arxiv.org/abs/2504.15895}, 
}

@misc{rlhf,
      title={Deep reinforcement learning from human preferences}, 
      author={Paul Christiano and Jan Leike and Tom B. Brown and Miljan Martic and Shane Legg and Dario Amodei},
      year={2023},
      eprint={1706.03741},
      archivePrefix={arXiv},
      primaryClass={stat.ML},
      url={https://arxiv.org/abs/1706.03741}, 
}

@misc{rlhf_distribution_1,
      title={Way Off-Policy Batch Deep Reinforcement Learning of Implicit Human Preferences in Dialog}, 
      author={Natasha Jaques and Asma Ghandeharioun and Judy Hanwen Shen and Craig Ferguson and Agata Lapedriza and Noah Jones and Shixiang Gu and Rosalind Picard},
      year={2019},
      eprint={1907.00456},
      archivePrefix={arXiv},
      primaryClass={cs.LG},
      url={https://arxiv.org/abs/1907.00456}, 
}

@misc{rlhf_distribution_2,
      title={Learning to summarize from human feedback}, 
      author={Nisan Stiennon and Long Ouyang and Jeff Wu and Daniel M. Ziegler and Ryan Lowe and Chelsea Voss and Alec Radford and Dario Amodei and Paul Christiano},
      year={2022},
      eprint={2009.01325},
      archivePrefix={arXiv},
      primaryClass={cs.CL},
      url={https://arxiv.org/abs/2009.01325}, 
}

@misc{gdpo,
      title={GDPO: Group reward-Decoupled Normalization Policy Optimization for Multi-reward RL Optimization}, 
      author={Shih-Yang Liu and Xin Dong and Ximing Lu and Shizhe Diao and Peter Belcak and Mingjie Liu and Min-Hung Chen and Hongxu Yin and Yu-Chiang Frank Wang and Kwang-Ting Cheng and Yejin Choi and Jan Kautz and Pavlo Molchanov},
      year={2026},
      eprint={2601.05242},
      archivePrefix={arXiv},
      primaryClass={cs.CL},
      url={https://arxiv.org/abs/2601.05242}, 
}

@misc{Group-Relative_Advantage_Biased,
      title={Your Group-Relative Advantage Is Biased}, 
      author={Fengkai Yang and Zherui Chen and Xiaohan Wang and Xiaodong Lu and Jiajun Chai and Guojun Yin and Wei Lin and Shuai Ma and Fuzhen Zhuang and Deqing Wang and Yaodong Yang and Jianxin Li and Yikun Ban},
      year={2026},
      eprint={2601.08521},
      archivePrefix={arXiv},
      primaryClass={cs.LG},
      url={https://arxiv.org/abs/2601.08521}, 
}

@misc{crst,
      title={Self-Training Elicits Concise Reasoning in Large Language Models}, 
      author={Tergel Munkhbat and Namgyu Ho and Seo Hyun Kim and Yongjin Yang and Yujin Kim and Se-Young Yun},
      year={2025},
      eprint={2502.20122},
      archivePrefix={arXiv},
      primaryClass={cs.CL},
      url={https://arxiv.org/abs/2502.20122}, 
}

@misc{vllm,
      title={Efficient Memory Management for Large Language Model Serving with PagedAttention}, 
      author={Woosuk Kwon and Zhuohan Li and Siyuan Zhuang and Ying Sheng and Lianmin Zheng and Cody Hao Yu and Joseph E. Gonzalez and Hao Zhang and Ion Stoica},
      year={2023},
      eprint={2309.06180},
      archivePrefix={arXiv},
      primaryClass={cs.LG},
      url={https://arxiv.org/abs/2309.06180}, 
}

@misc{pre_training,
      title={Megatron-LM: Training Multi-Billion Parameter Language Models Using Model Parallelism}, 
      author={Mohammad Shoeybi and Mostofa Patwary and Raul Puri and Patrick LeGresley and Jared Casper and Bryan Catanzaro},
      year={2020},
      eprint={1909.08053},
      archivePrefix={arXiv},
      primaryClass={cs.CL},
      url={https://arxiv.org/abs/1909.08053}, 
}

@misc{sample_polarity,
      title={Rethinking Sample Polarity in Reinforcement Learning with Verifiable Rewards}, 
      author={Xinyu Tang and Yuliang Zhan and Zhixun Li and Wayne Xin Zhao and Zhenduo Zhang and Zujie Wen and Zhiqiang Zhang and Jun Zhou},
      year={2025},
      eprint={2512.21625},
      archivePrefix={arXiv},
      primaryClass={cs.CL},
      url={https://arxiv.org/abs/2512.21625}, 
}

@book{coefficient_of_variation,
  title     = {The Cambridge Dictionary of Statistics},
  author    = {Everitt, Brian S. and Skrondal, Anders},
  year      = {2010},
  edition   = {4},
  publisher = {Cambridge University Press},
  address   = {Cambridge, UK}
}

@misc{kl_divergence,
  author       = {Schulman, John},
  title        = {Approximating KL Divergence},
  year         = {2020},
  howpublished = {\url{http://joschu.net/blog/kl-approx.html}}
}

@misc{revisit_grpo,
      title={Revisiting Group Relative Policy Optimization: Insights into On-Policy and Off-Policy Training}, 
      author={Youssef Mroueh and Nicolas Dupuis and Brian Belgodere and Apoorva Nitsure and Mattia Rigotti and Kristjan Greenewald and Jiri Navratil and Jerret Ross and Jesus Rios},
      year={2025},
      eprint={2505.22257},
      archivePrefix={arXiv},
      primaryClass={cs.LG},
      url={https://arxiv.org/abs/2505.22257}, 
}

@misc{advantage_importance,
      title={Kalman Filter Enhanced GRPO for Reinforcement Learning-Based Language Model Reasoning}, 
      author={Hu Wang and Congbo Ma and Ian Reid and Mohammad Yaqub},
      year={2025},
      eprint={2505.07527},
      archivePrefix={arXiv},
      primaryClass={cs.LG},
      url={https://arxiv.org/abs/2505.07527}, 
}

@misc{XLM-RoBERTa,
      title={Unsupervised Cross-lingual Representation Learning at Scale}, 
      author={Alexis Conneau and Kartikay Khandelwal and Naman Goyal and Vishrav Chaudhary and Guillaume Wenzek and Francisco Guzmán and Edouard Grave and Myle Ott and Luke Zettlemoyer and Veselin Stoyanov},
      year={2020},
      eprint={1911.02116},
      archivePrefix={arXiv},
      primaryClass={cs.CL},
      url={https://arxiv.org/abs/1911.02116}, 
}

@misc{LLMLingua-2,
      title={LLMLingua-2: Data Distillation for Efficient and Faithful Task-Agnostic Prompt Compression}, 
      author={Zhuoshi Pan and Qianhui Wu and Huiqiang Jiang and Menglin Xia and Xufang Luo and Jue Zhang and Qingwei Lin and Victor Rühle and Yuqing Yang and Chin-Yew Lin and H. Vicky Zhao and Lili Qiu and Dongmei Zhang},
      year={2024},
      eprint={2403.12968},
      archivePrefix={arXiv},
      primaryClass={cs.CL},
      url={https://arxiv.org/abs/2403.12968}, 
}

@article{sft_rl_connection1,
  title     = {Methods of Reducing Sample Size in Monte Carlo Computations},
  author    = {Kahn, H. and Marshall, A. W.},
  journal   = {Operations Research},
  volume    = {1},
  number    = {5},
  pages     = {263--278},
  year      = {1953},
  doi       = {10.1287/opre.1.5.263}
}

@article{sft_rl_connection2,
  title     = {Using Expectation-Maximization for Reinforcement Learning},
  author    = {Dayan, Peter and Hinton, Geoffrey E.},
  journal   = {Neural Computation},
  volume    = {9},
  number    = {2},
  pages     = {271--278},
  year      = {1997},
  doi       = {10.1162/neco.1997.9.2.271}
}

@inproceedings{important_sampling1,
 author = {Metelli, Alberto Maria and Papini, Matteo and Faccio, Francesco and Restelli, Marcello},
 booktitle = {Advances in Neural Information Processing Systems},
 editor = {S. Bengio and H. Wallach and H. Larochelle and K. Grauman and N. Cesa-Bianchi and R. Garnett},
 pages = {},
 publisher = {Curran Associates, Inc.},
 title = {Policy Optimization via Importance Sampling},
 url = {https://proceedings.neurips.cc/paper_files/paper/2018/file/6aed000af86a084f9cb0264161e29dd3-Paper.pdf},
 volume = {31},
 year = {2018}
}

@InProceedings{important_sampling2,
  title = 	 {Doubly Robust Off-policy Value Evaluation for Reinforcement Learning},
  author = 	 {Jiang, Nan and Li, Lihong},
  booktitle = 	 {Proceedings of The 33rd International Conference on Machine Learning},
  pages = 	 {652--661},
  year = 	 {2016},
  editor = 	 {Balcan, Maria Florina and Weinberger, Kilian Q.},
  volume = 	 {48},
  series = 	 {Proceedings of Machine Learning Research},
  address = 	 {New York, New York, USA},
  month = 	 {20--22 Jun},
  publisher =    {PMLR},
  url = 	 {https://proceedings.mlr.press/v48/jiang16.html}
}

@article{high_variance,
  title     = {On the Choice of Alternative Measures in Importance Sampling with Markov Chains},
  author    = {Andrad\'ottir, Sigr\'un and Heyman, Daniel P. and Ott, Teunis J.},
  journal   = {Operations Research},
  volume    = {43},
  number    = {3},
  pages     = {509--523},
  year      = {1995},
  doi       = {10.1287/opre.43.3.509}
}

@InProceedings{TRPO,
  title = 	 {Trust Region Policy Optimization},
  author = 	 {Schulman, John and Levine, Sergey and Abbeel, Pieter and Jordan, Michael and Moritz, Philipp},
  booktitle = 	 {Proceedings of the 32nd International Conference on Machine Learning},
  pages = 	 {1889--1897},
  year = 	 {2015},
  editor = 	 {Bach, Francis and Blei, David},
  volume = 	 {37},
  series = 	 {Proceedings of Machine Learning Research},
  address = 	 {Lille, France},
  month = 	 {07--09 Jul},
  publisher =    {PMLR},
  url = 	 {https://proceedings.mlr.press/v37/schulman15.html}
}

@misc{sft_is_rl,
      title={Supervised Fine Tuning on Curated Data is Reinforcement Learning (and can be improved)}, 
      author={Chongli Qin and Jost Tobias Springenberg},
      year={2025},
      eprint={2507.12856},
      archivePrefix={arXiv},
      primaryClass={cs.LG},
      url={https://arxiv.org/abs/2507.12856}, 
}

@misc{proximal_sft,
      title={Proximal Supervised Fine-Tuning}, 
      author={Wenhong Zhu and Ruobing Xie and Rui Wang and Xingwu Sun and Di Wang and Pengfei Liu},
      year={2025},
      eprint={2508.17784},
      archivePrefix={arXiv},
      primaryClass={cs.LG},
      url={https://arxiv.org/abs/2508.17784}, 
}

@misc{generalization_of_sft,
      title={On the Generalization of SFT: A Reinforcement Learning Perspective with Reward Rectification}, 
      author={Yongliang Wu and Yizhou Zhou and Zhou Ziheng and Yingzhe Peng and Xinyu Ye and Xinting Hu and Wenbo Zhu and Lu Qi and Ming-Hsuan Yang and Xu Yang},
      year={2025},
      eprint={2508.05629},
      archivePrefix={arXiv},
      primaryClass={cs.LG},
      url={https://arxiv.org/abs/2508.05629}, 
}

@misc{unified_view,
      title={Towards a Unified View of Large Language Model Post-Training}, 
      author={Xingtai Lv and Yuxin Zuo and Youbang Sun and Hongyi Liu and Yuntian Wei and Zhekai Chen and Xuekai Zhu and Kaiyan Zhang and Bingning Wang and Ning Ding and Bowen Zhou},
      year={2026},
      eprint={2509.04419},
      archivePrefix={arXiv},
      primaryClass={cs.LG},
      url={https://arxiv.org/abs/2509.04419}, 
}

@misc{entropy_early_exit,
      title={Entropy After $\langle \texttt{/Think} \rangle$ for reasoning model early exiting}, 
      author={Xi Wang and James McInerney and Lequn Wang and Nathan Kallus},
      year={2025},
      eprint={2509.26522},
      archivePrefix={arXiv},
      primaryClass={cs.LG},
      url={https://arxiv.org/abs/2509.26522}, 
}

@misc{thinking_intervention,
      title={Effectively Controlling Reasoning Models through Thinking Intervention}, 
      author={Tong Wu and Chong Xiang and Jiachen T. Wang and G. Edward Suh and Prateek Mittal},
      year={2025},
      eprint={2503.24370},
      archivePrefix={arXiv},
      primaryClass={cs.LG},
      url={https://arxiv.org/abs/2503.24370}, 
}

@misc{probing_hidden_state,
      title={Reasoning Models Know When They're Right: Probing Hidden States for Self-Verification}, 
      author={Anqi Zhang and Yulin Chen and Jane Pan and Chen Zhao and Aurojit Panda and Jinyang Li and He He},
      year={2025},
      eprint={2504.05419},
      archivePrefix={arXiv},
      primaryClass={cs.AI},
      url={https://arxiv.org/abs/2504.05419}, 
}

@misc{syncthink,
      title={SyncThink: A Training-Free Strategy to Align Inference Termination with Reasoning Saturation}, 
      author={Gengyang Li and Wang Cai and Yifeng Gao and Yunfang Wu},
      year={2026},
      eprint={2601.03649},
      archivePrefix={arXiv},
      primaryClass={cs.CL},
      url={https://arxiv.org/abs/2601.03649}, 
}

@misc{explorint_and_exploiting,
      title={Exploring and Exploiting the Inherent Efficiency within Large Reasoning Models for Self-Guided Efficiency Enhancement}, 
      author={Weixiang Zhao and Jiahe Guo and Yang Deng and Xingyu Sui and Yulin Hu and Yanyan Zhao and Wanxiang Che and Bing Qin and Tat-Seng Chua and Ting Liu},
      year={2025},
      eprint={2506.15647},
      archivePrefix={arXiv},
      primaryClass={cs.AI},
      url={https://arxiv.org/abs/2506.15647}, 
}

@misc{dart,
      title={DART: Difficulty-Adaptive Reasoning Truncation for Efficient Large Language Models}, 
      author={Ruofan Zhang and Bin Xia and Zhen Cheng and Cairen Jian and Minglun Yang and Ngai Wong and Yuan Cheng},
      year={2025},
      eprint={2511.01170},
      archivePrefix={arXiv},
      primaryClass={cs.AI},
      url={https://arxiv.org/abs/2511.01170}, 
}

@misc{greedy_decoding,
      title={Do LLMs Encode Functional Importance of Reasoning Tokens?}, 
      author={Janvijay Singh and Dilek Hakkani-Tür},
      year={2026},
      eprint={2601.03066},
      archivePrefix={arXiv},
      primaryClass={cs.CL},
      url={https://arxiv.org/abs/2601.03066}, 
}

@misc{efficient_reasoning_survey,
      title={Efficient Reasoning Models: A Survey}, 
      author={Sicheng Feng and Gongfan Fang and Xinyin Ma and Xinchao Wang},
      year={2025},
      eprint={2504.10903},
      archivePrefix={arXiv},
      primaryClass={cs.CL},
      url={https://arxiv.org/abs/2504.10903}, 
}

@misc{rl_via_self_distillation,
      title={Reinforcement Learning via Self-Distillation}, 
      author={Jonas Hübotter and Frederike Lübeck and Lejs Behric and Anton Baumann and Marco Bagatella and Daniel Marta and Ido Hakimi and Idan Shenfeld and Thomas Kleine Buening and Carlos Guestrin and Andreas Krause},
      year={2026},
      eprint={2601.20802},
      archivePrefix={arXiv},
      primaryClass={cs.LG},
      url={https://arxiv.org/abs/2601.20802}, 
}

@misc{self_distilled_reasoner,
      title={Self-Distilled Reasoner: On-Policy Self-Distillation for Large Language Models}, 
      author={Siyan Zhao and Zhihui Xie and Mengchen Liu and Jing Huang and Guan Pang and Feiyu Chen and Aditya Grover},
      year={2026},
      eprint={2601.18734},
      archivePrefix={arXiv},
      primaryClass={cs.LG},
      url={https://arxiv.org/abs/2601.18734}, 
}

@misc{self_distillation_enables_continual_learning,
      title={Self-Distillation Enables Continual Learning}, 
      author={Idan Shenfeld and Mehul Damani and Jonas Hübotter and Pulkit Agrawal},
      year={2026},
      eprint={2601.19897},
      archivePrefix={arXiv},
      primaryClass={cs.LG},
      url={https://arxiv.org/abs/2601.19897}, 
}

@article{on_policy_distillation,
  author = {Kevin Lu and Thinking Machines Lab},
  title = {On-Policy Distillation},
  journal = {Thinking Machines Lab: Connectionism},
  year = {2025},
  note = {https://thinkingmachines.ai/blog/on-policy-distillation},
  doi = {10.64434/tml.20251026},
}

@article{justrl,
  title={Justrl: Scaling a 1.5 b llm with a simple rl recipe},
  author={He, Bingxiang and Qu, Zekai and Liu, Zeyuan and Chen, Yinghao and Zuo, Yuxin and Qian, Cheng and Zhang, Kaiyan and Chen, Weize and Xiao, Chaojun and Cui, Ganqu and others},
  journal={arXiv preprint arXiv:2512.16649},
  year={2025}
}
\bibliographystyle{icml2026}

%%%%%%%%%%%%%%%%%%%%%%%%%%%%%%%%%%%%%%%%%%%%%%%%%%%%%%%%%%%%%%%%%%%%%%%%%%%%%%%
%%%%%%%%%%%%%%%%%%%%%%%%%%%%%%%%%%%%%%%%%%%%%%%%%%%%%%%%%%%%%%%%%%%%%%%%%%%%%%%
% APPENDIX
%%%%%%%%%%%%%%%%%%%%%%%%%%%%%%%%%%%%%%%%%%%%%%%%%%%%%%%%%%%%%%%%%%%%%%%%%%%%%%%
%%%%%%%%%%%%%%%%%%%%%%%%%%%%%%%%%%%%%%%%%%%%%%%%%%%%%%%%%%%%%%%%%%%%%%%%%%%%%%%
\newpage
\appendix
\onecolumn
\section{Related Work}
\label{Related Work}
\paragraph{Efficient reasoning}
Efficient reasoning methods can be broadly grouped into three categories: \textit{training-free methods, SFT-based methods, and RL-based methods}. 
(i) \textbf{Training-free} approaches aim to improve reasoning efficiency purely at inference time without additional training. Works in this line can be further categorized into prompt-based techniques, early-exit strategies, and activation steering. Prompt-based methods encourage models to produce more concise reasoning through explicit instructional prompts \citep{CoD,no_thinking,token_budget_prompt}. Early-exit strategies terminate the reasoning process based on carefully designed confidence or uncertainty metrics when further computation is deemed unnecessary \citep{deer,entropy_early_exit,syncthink}. Activation steering, by contrast, intervenes directly in the model’s internal activations to bias generation toward more efficient reasoning trajectories \citep{thinking_intervention,probing_hidden_state,explorint_and_exploiting}. 
(ii) \textbf{SFT-based} methods construct compressed reasoning datasets and fine-tune models on top of them. Concise Reasoning Self-Training (CRST) \citep{crst} fine-tunes models on self-generated concise reasoning paths obtained via best-of-$N$ sampling. TokenSkip \citep{tokenskip} prunes semantically unimportant tokens from CoT trajectories to produce compressed training data with varying compression ratios. StepEntropy \citep{step_entropy} replaces low-entropy steps in self-generated reasoning trajectories with a special \texttt{[skip]} token. DART \citep{dart} distills concise reasoning patterns from stronger models. \citet{greedy_decoding} proposes an iterative pruning approach that removes reasoning tokens whose removal minimally degrades the model likelihood under a specified objective.
(iii) \textbf{RL-based} methods optimize reasoning efficiency through policy optimization.
KIMI k1.5 \citep{KIMI} promotes shorter responses and penalizes longer responses among correct outputs, while explicitly penalizing long responses with incorrect answers.
ThinkPrune \citep{thinkprune} assigns a unit reward to responses that are both correct and within a predefined maximum length.
O1-Pruner \citep{O1-PRUNER} introduces a length-harmonizing reward defined as the ratio between the reasoning length of a reference model and that of the current policy.
L1 \citep{L1} trains models to satisfy user-specified length constraints.
ER-RL \citep{er_rl} assigns rewards only to correct responses, with shorter outputs receiving higher rewards.
LASER \citep{laser} employs a step-function reward based on a target length and dynamically adjusts this target during training. 
DLER \citep{dler_nvidia} introduces a training recipe that combines batch-wise reward normalization, higher clipping thresholds, dynamic sampling, and a simple truncation length penalty. 
\citet{xiaomi} proposes a mastery-gated, sample-level, soft RL compression scheme that penalizes long rollouts only when the model already solves the problem and has produced a shorter rollout.

\paragraph{Supervised fine-tuning and reinforcement learning}
The connection between supervised learning and reinforcement learning through importance weighting has deep theoretical roots \citep{sft_rl_connection1,sft_rl_connection2}. From the perspective of policy optimization, importance sampling provides a principled mechanism for leveraging data collected under one policy to optimize another by appropriately reweighting samples \citep{important_sampling1,important_sampling2}. A well-known challenge of this approach is that importance weights may become highly variable when the behavior and target policies differ substantially \citep{high_variance}. To control this instability, trust-region methods \citep{TRPO} and proximal policy optimization \citep{ppo} explicitly restrict the magnitude of policy updates relative to a reference policy. Motivated by these ideas, Recent work on LLM fine-tuning has proposed a range of techniques, including importance-weighted supervised objectives \citep{sft_is_rl}, proximal variants of supervised fine-tuning \citep{proximal_sft}, probability-based reweighting schemes in dynamic fine-tuning \citep{generalization_of_sft}, and unified perspectives that reconcile supervised and reinforcement learning paradigms \citep{unified_view}. In contrast to prior approaches, our method anchors the auxiliary distribution directly to the base model, extending importance-weighted formulations with an explicit stabilization mechanism while maintaining their theoretical guarantees. In contrast to these approaches, we present a new perspective that places SFT in an on-policy data regime and establishes a direct connection between SFT and RL.

\section{Unified Reward Formulations}
\label{app:Unified Reward Formulations}
\begin{figure*}[h!]
\begin{tcolorbox}[
  colback=gray!10,
  colframe=black,
  title=\textbf{Prompt Template},
  fontupper=\ttfamily,
  breakable
]
<|begin\_of\_sentence|><|User|>\{question\}
Please reason step by step, and put your final answer within \textbackslash\textbackslash boxed\{\}.

\vspace{0.5em}

<|Assistant|><think>
\end{tcolorbox}
\caption{Prompt template used for both training and evaluation.}
\label{fig:prompt_template}
\end{figure*}
\begin{figure*}[h!]
\begin{tcolorbox}[
  colback=gray!10,
  colframe=black,
  title=\textbf{CoD Prompt Template},
  fontupper=\ttfamily,
  breakable
]
<|begin\_of\_sentence|><|User|>\{question\}
Please reason step by step, but only keep a minimum draft for each thinking step, with 5 words at most, and put your final answer within \textbackslash\textbackslash boxed\{\}.

\vspace{0.5em}

<|Assistant|><think>
\end{tcolorbox}
\caption{Prompt template used for CoD.}
\label{fig:CoD_template}
\end{figure*}
As discussed in \autoref{Reward Shaping for Efficient Reasoning}, the rewards used in existing
RL-based efficient reasoning methods can be formalized in a unified form:
\begin{equation}
R_{\text{Eff}}(o_i \mid q)
=
R_{\text{Acc}}(o_i \mid q)
+
\gamma(o_i \mid q)\, R_{\text{Len}}(o_i \mid q),
\end{equation}

Below we summarize how representative prior methods instantiate
$R_{\text{Acc}}(o_i \mid q)$, $\gamma(o_i \mid q)$, and $R_{\text{Len}}(o_i \mid q)$.
We use $L(o_i)$ to denote the generated token length of $o_i$.

% \paragraph{Vanilla Truncation~\citep{thinkprune,dler_nvidia}}
% \begin{align}
% R_{\text{Acc}}(o_i \mid q) &= 0, \\
% \gamma(o_i \mid q) &= 1, \\
% R_{\text{Len}}(o_i \mid q) &=
% \begin{cases}
% R_{\text{Acc}}(o_i \mid q), & L(o_i) \le L_T, \\
% \rho, & L(o_i) > L_T,
% \end{cases}
% \end{align}
% where $L_T$ denotes a predefined fixed maximum generation length,
% and $\rho$ is typically set to zero.

% \paragraph{ThinkPrune~\citep{thinkprune}}
% \begin{align}
% R_{\text{Acc}}(o_i \mid q) &= 0, \\
% \gamma(o_i \mid q) &= 1, \\
% R_{\text{Len}}(o_i \mid q) &=
% \begin{cases}
% R_{\text{Acc}}(o_i \mid q), & L(o_i) \le L_A, \\
% \rho, & L(o_i) > L_A,
% \end{cases}
% \end{align}
% where $L_A$ denotes a dynamically adjusted maximum generation length,
% which is set to 4k initially and then reduced to 3k and 2k.

\paragraph{ER-RL~\citep{er_rl}}
\begin{align}
R_{\text{Acc}}(o_i \mid q) &= R_i, \\
\gamma(o_i \mid q) &= \mathbb{I}\big(R_i = 1\big), \\
R_{\text{Len}}(o_i \mid q) &=
- \alpha \cdot \sigma\!\left(
\frac{L(o_i) - \operatorname{Mean}(L)}{\operatorname{Std}(L)}
\right),
\end{align}
where $R_i \in \{0,1\}$ is a binary correctness reward,
$\alpha \in [0,1)$ is a scaling coefficient,
and $\sigma(\cdot)$ denotes the sigmoid function.

\paragraph{Kimi-k1.5~\citep{KIMI}}
\begin{align}
R_{\text{Acc}}(o_i \mid q) &= R_i, \\
\gamma(o_i \mid q) &= 1, \\
R_{\text{Len}}(o_i \mid q) &=
\begin{cases}
0.5 - \dfrac{L(o_i) - L_{\min}}{L_{\max} - L_{\min}}, &
\mathbb{I}(R_i = 1), \\[6pt]
\min\!\left(0,\,
0.5 - \dfrac{L(o_i) - L_{\min}}{L_{\max} - L_{\min}}\right), &
\mathbb{I}(R_i = 0),
\end{cases}
\end{align}
where $L_{\min} = \min_i L(o_i)$ and $L_{\max} = \max_i L(o_i)$.

\paragraph{L1-Exact \citep{L1}}
\begin{align}
R_{\text{Acc}}(o_i \mid q) &= R_i, \\
\gamma(o_i \mid q) &= 1, \\
R_{\text{Len}}(o_i \mid q) &= -\alpha \cdot \lvert L(o_i) - L_T \rvert,
\end{align}
where $L_T$ is a user-specific target length.

\paragraph{L1-Max~\citep{L1}}
\begin{align}
R_{\text{Acc}}(o_i \mid q) &= 0, \\
\gamma(o_i \mid q) &= \mathbb{I}\big(R_i = 1\big), \\
R_{\text{Len}}(o_i \mid q) &=
\operatorname{clip}\!\left(
\alpha \cdot (L(o_i) - L_T) + \delta,\ 0,\ 1
\right),
\end{align}
where $L_T$ is a user-specific maximum length constraint.

\paragraph{LASER-DE~\citep{laser}}
\begin{align}
R_{\text{Acc}}(o_i \mid q) &= R_i, \\
\gamma(o_i \mid q) &= 1, \\
R_{\text{Len}}(o_i \mid q)
&=
\alpha\, \mathbb{I}(R_i = 1)\mathbb{I}(L(o_i) \le L_A)
+
\alpha\, \mathbb{I}(R_i = 0)\mathbb{I}(L(o_i) > L_A),
\end{align}
where $L_A$ denotes a dynamically adjusted length threshold.

\paragraph{Mastery-Gated Length Penalty~\citep{xiaomi}}
\begin{align}
R_{\text{Acc}}(o_i \mid q) &= R_i, \\
\gamma(o_i \mid q) &= \mathbb{I}\big(\hat{p}(q) = 1\big), \\
R_{\text{Len}}(o_i \mid q) &=
\begin{cases}
0, & L(o_i) \le L_{\text{start}}(q), \\[4pt]
-1, & L(o_i) > L_{\max}(q), \\[6pt]
-\dfrac{L(o_i)-L_{\text{start}}(q)}{L_{\max}(q)-L_{\text{start}}(q)}, &
L_{\text{start}}(q) < L(o_i) \le L_{\max}(q),
\end{cases}
\end{align}
where $R_i \in \{0,1\}$ is a binary correctness reward,
$\hat{p}(q) = \frac{1}{N}\sum_{i=1}^{N} R_i$ denotes the correctness rate
over $N$ rollouts for question $q$,
$L_{\text{start}}(q)$ is the median generated token length among correct rollouts,
and $L_{\max}(q)$ is the maximum generated token length among correct rollouts.

\section{Derivation of the GRPO Gradient}
\label{app:grpo_grad}

We derive the gradient of the GRPO objective used in the main text.
We consider $(q,a)\sim\mathcal{D}$ and $G$ rollouts $\{o_i\}_{i=1}^G$ sampled from a fixed
rollout snapshot $\pi_{\theta_{\text{old}}}(\cdot\mid q)$, where
$o_i=(o_{i,1},\ldots,o_{i,|o_i|})$.

\paragraph{GRPO objective.}
The GRPO objective is
\begin{equation}
\label{eq:grpo_obj}
\begin{aligned}
J_{\text{GRPO}}(\theta)
=
\mathbb{E}_{(q,a)\sim\mathcal{D},\,\{o_i\}_{i=1}^{G}\sim\pi_{\theta_{\text{old}}}(\cdot\mid q)}
\Bigg[
\frac{1}{G}\sum_{i=1}^{G}\frac{1}{|o_i|}
\sum_{t=1}^{|o_i|}
\min\Big(
r_{i,t}(\theta)\hat{A}_{i,t},\;
\mathrm{clip}(r_{i,t}(\theta),1-\epsilon,1+\epsilon)\hat{A}_{i,t}
\Big)
- \\
\beta\,D_{\mathrm{KL}}(\pi_\theta \,\|\, \pi_{\mathrm{ref}})
\Bigg],
\end{aligned}
\end{equation}
where
$r_{i,t}(\theta)=\pi_\theta(o_{i,t}\mid q,o_{i,<t})/
\pi_{\theta_{\text{old}}}(o_{i,t}\mid q,o_{i,<t})$.
The KL term is estimated using the following unbiased per-token estimator
\citep{kl_divergence}:
\begin{equation}
\label{eq:kl_est}
D_{\mathrm{KL},\,i,t}(\theta)
=
\frac{\pi_{\mathrm{ref}}(o_{i,t}\mid o_{i,<t})}{\pi_\theta(o_{i,t}\mid q,o_{i,<t})}
-
\log\frac{\pi_{\mathrm{ref}}(o_{i,t}\mid o_{i,<t})}{\pi_\theta(o_{i,t}\mid q,o_{i,<t})}
-1.
\end{equation}

\paragraph{Step 1: Differentiate the original objective.}
Since rollouts are sampled from the fixed snapshot $\pi_{\theta_{\text{old}}}$, the sampling
distribution does not depend on $\theta$, so we can move $\nabla_\theta$ inside the expectation:
\begin{equation}
\label{eq:move_grad_in}
\nabla_\theta J_{\text{GRPO}}(\theta)
=
\mathbb{E}\Bigg[
\frac{1}{G}\sum_{i=1}^{G}\frac{1}{|o_i|}
\sum_{t=1}^{|o_i|}
\nabla_\theta \,\ell_{i,t}(\theta)
\Bigg],
\end{equation}
where
\begin{equation}
\label{eq:ell_def}
\ell_{i,t}(\theta)
\triangleq
\min\Big(
r_{i,t}(\theta)\hat{A}_{i,t},\;
\mathrm{clip}(r_{i,t}(\theta),1-\epsilon,1+\epsilon)\hat{A}_{i,t}
\Big)
-
\beta\,D_{\mathrm{KL},\,i,t}(\theta).
\end{equation}

\paragraph{Step 2: Simplification under the single-update assumption.}
To simplify the analysis, we consider a single policy update after each exploration stage,
so that $\pi_{\theta_{\text{old}}} = \pi_\theta$. In this case, the PPO-style $\min(\cdot)$
and clipping operations can be removed:
\begin{equation}
\label{eq:ppo_clip_inactive}
\min\Big(
r_{i,t}(\theta)\hat{A}_{i,t},\;
\mathrm{clip}(r_{i,t}(\theta),1-\epsilon,1+\epsilon)\hat{A}_{i,t}
\Big)
=
r_{i,t}(\theta)\hat{A}_{i,t}.
\end{equation}

Taking the gradient of $r_{i,t}(\theta)\hat{A}_{i,t}$ with respect to $\theta$, we obtain
\begin{equation}
\label{eq:grad_rA}
\nabla_\theta\big(r_{i,t}(\theta)\hat{A}_{i,t}\big)
=
\hat{A}_{i,t}\nabla_\theta r_{i,t}(\theta)
=
\hat{A}_{i,t}\nabla_\theta \log \pi_\theta(o_{i,t}\mid q,o_{i,<t}),
\end{equation}
where
\begin{equation}
\label{eq:grad_ratio}
\begin{aligned}
\nabla_\theta r_{i,t}(\theta)
&=
\nabla_\theta
\frac{\pi_\theta(o_{i,t}\mid q,o_{i,<t})}
{\pi_{\theta_{\text{old}}}(o_{i,t}\mid q,o_{i,<t})} \\
&=
\frac{1}{\pi_{\theta_{\text{old}}}(o_{i,t}\mid q,o_{i,<t})}
\nabla_\theta \pi_\theta(o_{i,t}\mid q,o_{i,<t}) \\
&=
\frac{\pi_\theta(o_{i,t}\mid q,o_{i,<t})}
{\pi_{\theta_{\text{old}}}(o_{i,t}\mid q,o_{i,<t})}
\nabla_\theta \log \pi_\theta(o_{i,t}\mid q,o_{i,<t}) \\
&=
r_{i,t}(\theta)\nabla_\theta \log \pi_\theta(o_{i,t}\mid q,o_{i,<t}) \\
&=
\nabla_\theta \log \pi_\theta(o_{i,t}\mid q,o_{i,<t}),
\end{aligned}
\end{equation}
where the last equality uses $\pi_{\theta_{\text{old}}}=\pi_\theta$, hence $r_{i,t}(\theta)=1$.

\paragraph{Step 3: Gradient of the KL estimator.}
Taking the gradient of $D_{\mathrm{KL},\,i,t}(\theta)$ in \eqref{eq:kl_est}, we obtain
\begin{equation}
\label{eq:grad_kl}
\begin{aligned}
\nabla_\theta D_{\mathrm{KL},\,i,t}(\theta)
&=
\nabla_\theta\!\left(
\frac{\pi_{\mathrm{ref}}(o_{i,t}\mid o_{i,<t})}{\pi_\theta(o_{i,t}\mid q,o_{i,<t})}
-
\log\frac{\pi_{\mathrm{ref}}(o_{i,t}\mid o_{i,<t})}{\pi_\theta(o_{i,t}\mid q,o_{i,<t})}
-1
\right) \\
&=
\nabla_\theta\!\left(
\frac{\pi_{\mathrm{ref}}(o_{i,t}\mid o_{i,<t})}{\pi_\theta(o_{i,t}\mid q,o_{i,<t})}
+
\log \pi_\theta(o_{i,t}\mid q,o_{i,<t})
\right) \\
&=
-\frac{\pi_{\mathrm{ref}}(o_{i,t}\mid o_{i,<t})}{\pi_\theta(o_{i,t}\mid q,o_{i,<t})^2}
\nabla_\theta \pi_\theta(o_{i,t}\mid q,o_{i,<t})
+
\nabla_\theta \log \pi_\theta(o_{i,t}\mid q,o_{i,<t}) \\
&=
-\frac{\pi_{\mathrm{ref}}(o_{i,t}\mid o_{i,<t})}{\pi_\theta(o_{i,t}\mid q,o_{i,<t})^2}
\pi_\theta(o_{i,t}\mid q,o_{i,<t})
\nabla_\theta \log \pi_\theta(o_{i,t}\mid q,o_{i,<t})
+
\nabla_\theta \log \pi_\theta(o_{i,t}\mid q,o_{i,<t}) \\
&=
-\left(
\frac{\pi_{\mathrm{ref}}(o_{i,t}\mid o_{i,<t})}{\pi_\theta(o_{i,t}\mid q,o_{i,<t})}
-1
\right)
\nabla_\theta \log \pi_\theta(o_{i,t}\mid q,o_{i,<t}).
\end{aligned}
\end{equation}

\paragraph{Step 4: Combine terms.}
Combining \eqref{eq:grad_rA} and \eqref{eq:grad_kl} in $\nabla_\theta \ell_{i,t}(\theta)
= \nabla_\theta\big(r_{i,t}(\theta)\hat{A}_{i,t}\big) - \beta \nabla_\theta D_{\mathrm{KL},\,i,t}(\theta)$,
we obtain
\begin{equation}
\label{eq:grad_ell}
\begin{aligned}
\nabla_\theta \ell_{i,t}(\theta)
&=
\hat{A}_{i,t}\nabla_\theta \log \pi_\theta(o_{i,t}\mid q,o_{i,<t})
-\beta\!\left(
-\Big(
\frac{\pi_{\mathrm{ref}}(o_{i,t}\mid o_{i,<t})}{\pi_\theta(o_{i,t}\mid q,o_{i,<t})}
-1
\Big)
\nabla_\theta \log \pi_\theta(o_{i,t}\mid q,o_{i,<t})
\right) \\
&=
\Big(
\hat{A}_{i,t}
+
\beta\Big(
\frac{\pi_{\mathrm{ref}}(o_{i,t}\mid o_{i,<t})}{\pi_\theta(o_{i,t}\mid q,o_{i,<t})}
-1
\Big)
\Big)
\nabla_\theta \log \pi_\theta(o_{i,t}\mid q,o_{i,<t}).
\end{aligned}
\end{equation}
Substituting \eqref{eq:grad_ell} into \eqref{eq:move_grad_in} yields
\begin{equation}
\begin{aligned}
\nabla_\theta J_{\text{GRPO}}(\theta)
=
\mathbb{E}_{(q,a)\sim\mathcal{D},\,\{o_i\}_{i=1}^{G}\sim\pi_{\theta_{\text{old}}}(\cdot\mid q)}
\Bigg[
\frac{1}{G}\sum_{i=1}^{G}\frac{1}{|o_i|}
\sum_{t=1}^{|o_i|}
\Big(
\hat{A}_{i,t}
+
\beta\Big(
\frac{\pi_{\mathrm{ref}}(o_{i,t}\mid o_{i,<t})}
{\pi_\theta(o_{i,t}\mid q,o_{i,<t})}
- 1
\Big)
\Big)
\nabla_\theta \log \pi_\theta(o_{i,t}\mid q,o_{i,<t})
\Bigg],
\end{aligned}
\end{equation}
which matches the gradient expression reported in the main text.

\section{REINFORCE Algorithm}
\label{app:reinforce}

REINFORCE \cite{REINFORCEMENT} is a Monte Carlo policy gradient method that directly optimizes a parameterized stochastic policy $\pi_\theta(a \mid s)$. We consider an episodic RL setting in which a trajectory $\tau = (s_0, a_0, r_0, s_1, a_1, r_1, \dots, s_T)$ is generated by following $\pi_\theta$. The discounted return from time step $t$ is defined as $G_t = \sum_{k=t}^{T-1} \gamma^{k-t} r_k$, where $\gamma \in [0,1]$ is the discount factor. The objective is to maximize the expected cumulative return
\[
J(\theta) = \mathbb{E}_{\tau \sim \pi_\theta} \left[ \sum_{t=0}^{T-1} \gamma^t r_t \right].
\]

Using the log-derivative trick, the gradient of the objective can be written as
\[
\nabla_\theta J(\theta)
= \mathbb{E}_{\tau \sim \pi_\theta} \left[ \sum_{t=0}^{T-1} \nabla_\theta \log \pi_\theta(a_t \mid s_t)\, G_t \right].
\]
In practice, this expectation is approximated using Monte Carlo samples from complete episodes, yielding the update rule
\[
\theta \leftarrow \theta + \alpha \sum_{t=0}^{T-1} \nabla_\theta \log \pi_\theta(a_t \mid s_t)\, G_t ,
\]
where $\alpha$ denotes the learning rate.

\section{On-Policy Supervised Fine-Tuning}
\label{app:onpolicy_sft}
This section presents the complete training procedure for on-policy SFT, as summarized in
Algorithm~\ref{alg:onpolicy-sft}.

\section{Experimental Setup Details}
\label{app:Experimental Setup Details}
\paragraph{Dataset Details} DeepScaleR \citep{deepscaler2025} is a large-scale collection of mathematical reasoning problems aggregated from multiple established sources, including AIME (1983–2023), AMC, Omni-Math \citep{omni}, and STILL \citep{STILL}.

\paragraph{Training Details} During training, we use a group size of 8 and a global batch size of 64, and train for a single epoch. We adopt the AdamW optimizer with a learning rate of \(1 \times 10^{-7}\).

\paragraph{Evaluation Details}
We evaluate the effectiveness of On-Policy SFT on five widely used mathematical reasoning benchmarks: GSM8K \citep{gsm8k}, MATH-500 \citep{math500}, AMC23 \citep{AMC2023}, AIME24 \citep{AIME2024}, and AIME25 \citep{AIME2025}. Following the recommendations of \citet{DeepSeek-R1}, we adopt decoding hyperparameters with temperature set to 0.6 and top-p set to 0.95. The maximum number of output tokens is fixed at 32,768. For each question, we sample the model outputs \(N\) times to support multi-sample evaluation. The value of \(N\) is set to 64 for AMC23, AIME24, and AIME25, and to 16 for the remaining datasets.
All inference is conducted on a single A800 GPU using the vLLM library \citep{vllm}.
Details of the prompt templates are provided in Appendix~\ref{app:prompt_ours}.

\paragraph{Metrics}
We report five evaluation metrics: accuracy (\textbf{Acc}, in \%), \textbf{Pass@\(N\)} (in \%), average response token length (\textbf{Tok}), compression rate (\textbf{CR}, in \%), and token efficiency (\textbf{Eff}). 
Acc is defined as the proportion of correct responses among all sampled model outputs in the evaluation set. 
Pass@\(N\) measures the fraction of problems for which at least one correct solution is produced among \(N\) independently sampled outputs, where \(N\) corresponds to the per-problem sampling budget used during evaluation. We report Pass@\(N\) to assess whether the training procedure preserves the model’s ability to explore diverse solution paths across multiple sampling attempts. 
Tok measures the average number of generated tokens per sampled response. 
CR is defined as the ratio between the average number of tokens in the generated response and that of the original CoT.
Eff is defined as the ratio of accuracy to average response length (Eff = Acc / Tok, in \%), serving as an indicator of the correctness and reasoning efficiency trade-off.

\section{Prompt Templates}
\label{Prompt Templates}
\subsection{On-Policy SFT Prompt Template}
\label{app:prompt_ours}
The prompt template is illustrated in \autoref{fig:prompt_template}, which follows the design proposed in DeepSeek-R1~\citep{DeepSeek-R1}. This prompt template is used by all methods in our experiments except for CoD. For these methods, identical prompt templates are used for both training and evaluation, thereby avoiding potential distribution shifts caused by prompt mismatches and ensuring that performance gains can be attributed to the proposed On-Policy SFT procedure rather than prompt engineering.

\subsection{Chain-of-Draft (CoD) Prompt Template}
\label{app:prompt_cod}
We follow the prompt template proposed in Chain-of-Draft~\citep{CoD}. The prompt is shown in \autoref{fig:CoD_template}.

\section{Baseline Implementation Details}
\label{TokenSkip Implementation Details}
\paragraph{TokenSkip}
Due to the limited context length of XLM-RoBERTa~\citep{XLM-RoBERTa}, which serves as the backbone of LLMLingua-2~\citep{LLMLingua-2}, the model only supports a maximum context window of 512 tokens. In our implementation, self-generated CoT sequences are first split into non-overlapping chunks of 512 tokens. LLMLingua is then applied independently to each chunk to identify semantically unimportant tokens, which are subsequently removed to construct the training dataset. After training, we evaluate compression ratios ranging from 0.1 to 0.9 in increments of 0.1, and select 0.7 as it provides the best performance--efficiency trade-off.

\paragraph{L1}
we report the \emph{L1-max} variant, which requires the output to be no longer than the specified target length, with the target length set to 3{,}500.

\begin{algorithm}[t]
\caption{On-Policy Supervised Fine-Tuning}
\label{alg:onpolicy-sft}
\begin{algorithmic}[1]
\Require Training set $\mathcal{D}=\{(q,a)\}$; initial policy $\pi_{\theta}$;
rollouts per prompt $G$; length limit $L$; batch size $B$;
learning rate $\eta$; total steps $T$.
\Ensure Updated policy parameters $\theta$.

\For{$k=1$ to $T$}
    \State Sample a minibatch $\mathcal{B}=\{(q_b,a_b)\}_{b=1}^{B}$ from $\mathcal{D}$.
    \State Set $\theta_{\text{old}} \leftarrow \theta$. \Comment{on-policy snapshot}
    \State Initialize $\mathcal{C}_L \leftarrow \emptyset$ and $M \leftarrow 1$.
    \Comment{valid rollouts and max valid length}
    \For{$b=1$ to $B$}
        \For{$i=1$ to $G$}
            \State Sample a rollout $o_{b,i} \sim \pi_{\theta_{\text{old}}}(\cdot \mid q_b)$.
            \If{$\textsc{Correct}(o_{b,i}, a_b)\ \wedge\ |o_{b,i}|\le L$}
                \State Add $(q_b,o_{b,i})$ to $\mathcal{C}_L$.
                \State $M \leftarrow \max(M, |o_{b,i}|)$.
            \EndIf
        \EndFor
    \EndFor

    \If{$|\mathcal{C}_L|>0$}
        \State Compute the on-policy SFT objective over $\mathcal{C}_L$:
        \State \hspace{1.2em}
        $
        J(\theta)
        =
        \frac{1}{BG}
        \sum_{(q,o)\in\mathcal{C}_L}
        \frac{1}{M}
        \sum_{t=1}^{|o|}
        \log \pi_{\theta}(o_t \mid q, o_{<t})
        $.
        \State Compute loss $\mathcal{L}(\theta)\leftarrow -J(\theta)$.
        \State Update $\theta \leftarrow \theta - \eta \nabla_{\theta}\mathcal{L}(\theta)$.
    \EndIf
\EndFor
\end{algorithmic}
\end{algorithm}

\section{coefficient of variation}
\label{coefficient of variation}
Formally, for a given question $q$, we generate $N$ responses, where $N$ is set identically to the evaluation setup described in \autoref{Experimental Setup}. Let $\ell_{qj}$ denote the token length of the $j$-th response. The coefficient of variation of generation lengths for question $q$ is given by
\begin{equation}
\mathrm{NormStd}_q
=
\frac{
\sqrt{\frac{1}{N} \sum_{j=1}^{N} \left( \ell_{qj} - \frac{1}{N} \sum_{j=1}^{N} \ell_{qj} \right)^2}
}{
\frac{1}{N} \sum_{j=1}^{N} \ell_{qj}
}.
\end{equation}
\section{On-Policy Requirement for Rollout Temperature}
\label{On-Policy Requirement for Rollout Temperature}
Let \(z_{\theta_{\text{old}}}(y \mid q, o_{<t})\) denote the pre-softmax logit assigned by the
model with parameters \(\theta_{\text{old}}\) to token \(y\), given the question \(q\) and
the previously generated prefix \(o_{<t}\).
When rollouts are generated using temperature-scaled sampling, the token-level behavior
policy at timestep \(t\) is defined as
\begin{equation}
\pi^{(T)}_{\theta_{\text{old}}}\!\left(o_{t} \mid q, o_{<t}\right)
=
\frac{\exp\!\left(z_{\theta_{\text{old}}}\!\left(o_{t} \mid q, o_{<t}\right)/T\right)}
{\sum_{y}\exp\!\left(z_{\theta_{\text{old}}}\!\left(y \mid q, o_{<t}\right)/T\right)}.
\end{equation}
This token-level policy induces a trajectory-level distribution over responses
\(o = (o_1,\ldots,o_{|o|})\) according to
\begin{equation}
\Pr_T(o \mid q)
=
\prod_{t=1}^{|o|}
\pi^{(T)}_{\theta_{\text{old}}}\!\left(o_{t} \mid q, o_{<t}\right).
\end{equation}
Therefore, when rollouts are sampled with temperature \(T\), the expectation in
\autoref{eq:on_policy_sft} is taken with respect to the distribution \(\Pr_T(\cdot \mid q)\),
rather than \(\pi_{\theta_{\text{old}}}(\cdot \mid q)\).

When \(T = 1\), the temperature-scaled token distribution reduces to the original model
policy, i.e.,
\begin{equation}
\pi^{(1)}_{\theta_{\text{old}}}\!\left(o_t \mid q, o_{<t}\right)
=
\pi_{\theta_{\text{old}}}\!\left(o_t \mid q, o_{<t}\right),
\end{equation}
and consequently \(\Pr_1(o \mid q) = \pi_{\theta_{\text{old}}}(o \mid q)\).
In contrast, for any \(T \neq 1\),
\begin{equation}
\pi^{(T)}_{\theta_{\text{old}}}\!\left(\cdot \mid q, o_{<t}\right)
\neq
\pi_{\theta_{\text{old}}}\!\left(\cdot \mid q, o_{<t}\right),
\end{equation}
which implies \(\Pr_T(\cdot \mid q) \neq \pi_{\theta_{\text{old}}}(\cdot \mid q)\).
Hence, when \(T \neq 1\), rollouts are generated by a behavior policy that differs from
\(\pi_{\theta_{\text{old}}}\), which directly leads to an off-policy optimization
objective in \autoref{eq:on_policy_sft}.
\begin{figure}[t]
    \centering
    \includegraphics[width=0.8\linewidth]{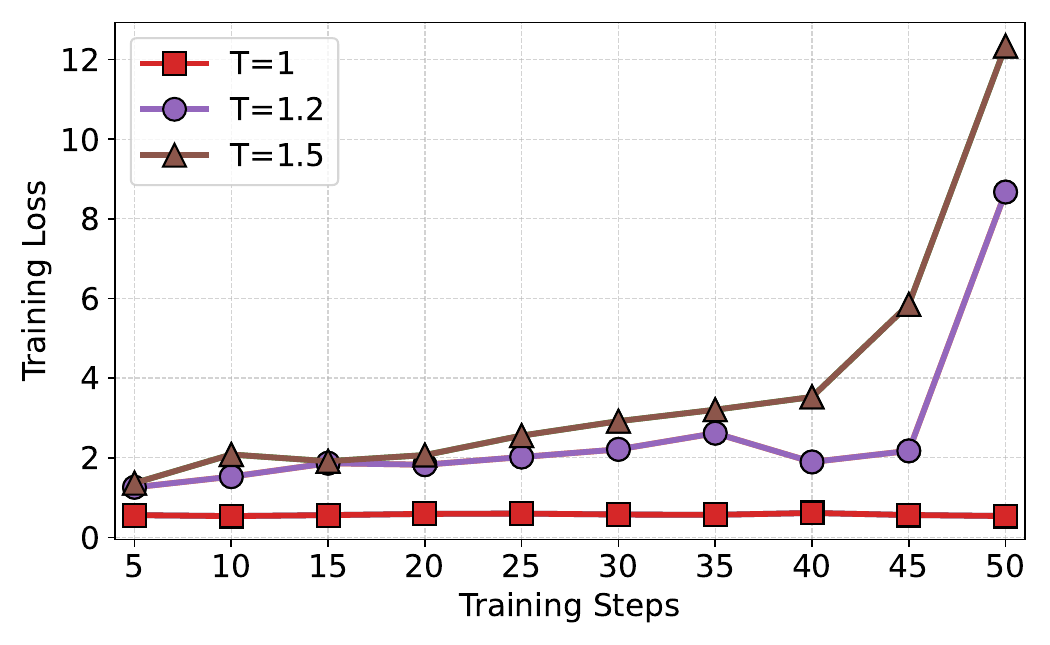}
    \caption{Training loss curves of the 1.5B model on MATH-500 under different rollout temperatures (1.0, 1.2, and 1.5).}
    \label{fig:training_loss_temp}
\end{figure}

\section{A Diagnostic Tool for Identifying Inefficient Tokens}
\label{sec:kl_diagnostic}
\begin{figure}[t]
    \centering
    \includegraphics[width=\linewidth]{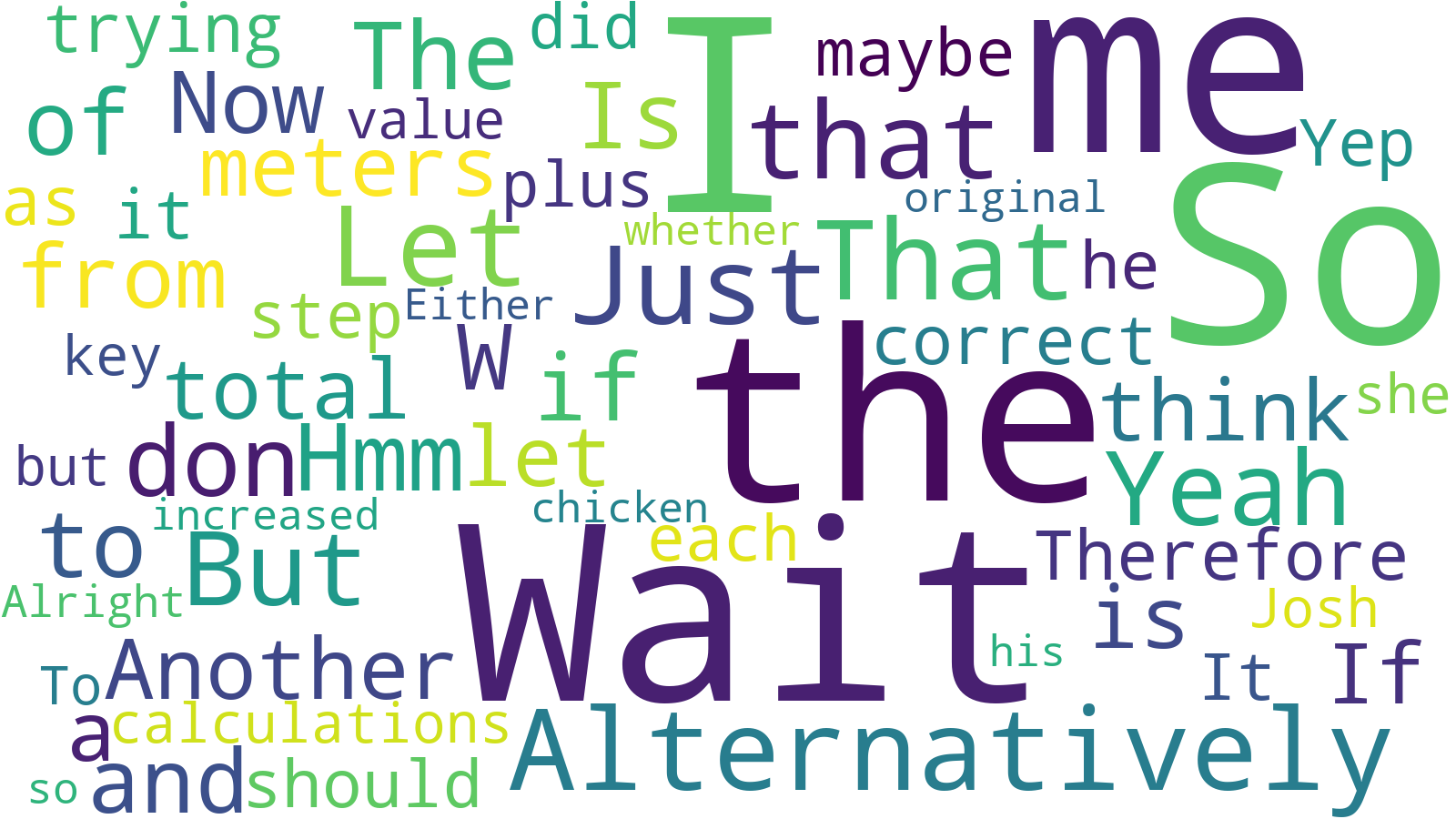}
    \caption{Next-token word cloud of the original model at positions with large KL divergence from the on-policy SFT–trained efficient model. Larger font sizes indicate greater divergence.}
    \label{fig:wordcloud}
\end{figure}
Inspired by recent work on on-policy distillation \citep{on_policy_distillation,self_distillation_enables_continual_learning, self_distilled_reasoner, rl_via_self_distillation}, we introduce a principled diagnostic tool for identifying which tokens in an original CoT are most likely to steer the model toward inefficient reasoning trajectories. Specifically, we first sample an original CoT from the base model, and then feed the same prefix (the prompt together with the sampled CoT prefix) into the efficient model obtained after on-policy SFT via teacher forcing. At each position, we measure the divergence between the next-token distributions of the original and efficient models conditioned on the same history.
Let $x_{1:t}$ denote the shared prefix at position $t$. We denote the next-token distributions of the original and efficient models as
$p_{\text{orig}}(\cdot \mid x_{1:t})$ and $p_{\text{eff}}(\cdot \mid x_{1:t})$, respectively. The token-level divergence is computed as
\begin{equation}
\begin{aligned}
&D_t
\;=\;
\mathrm{KL}\!\left(p_{\text{orig}}(\cdot \mid x_{1:t}) \,\|\, p_{\text{eff}}(\cdot \mid x_{1:t})\right)
\;=\;
\sum_{v \in \mathcal{V}}\\
& p_{\text{orig}}(v \mid x_{1:t})
\Bigl(\log p_{\text{orig}}(v \mid x_{1:t}) - \log p_{\text{eff}}(v \mid x_{1:t})\Bigr).    
\end{aligned}
\end{equation}
A larger $D_t$ indicates a stronger disagreement between the two models regarding the next-token distribution, thereby pinpointing positions at which the efficient model diverges most sharply from the original trajectory.

We visualize the next-token word cloud of the original model at positions with large KL divergence under a fixed prefix in \autoref{fig:wordcloud}, where larger font sizes indicate greater disagreement with the efficient model. We observe that tokens such as ``Wait'', ``Alternatively'', and ``Hmm'', which often signal hesitation or shifts in reasoning, are consistently identified as inefficient. This suggests that excessive reflection or backtracking is not always required, and that a more confident reasoning trajectory may yield correct solutions with substantially lower latency.
In addition, we uncover another intriguing pattern: first-person singular pronouns such as ``I'' and ``me'' also exhibit large divergence. In contrast, the efficient variant tends to prefer plural forms such as ``we'' and ``us.'' We hypothesize that this behavior may be related to pretraining data distributions, where corpora involving collective expressions are often more concise, potentially offering a new perspective on achieving efficient reasoning. We include several question-level diagnostic examples in the Appendix~\ref{appendix:kl_examples}.

\section{Qualitative Examples of KL-Based Diagnostics}
\label{appendix:kl_examples}
\begin{figure}[t]
    \centering
    \includegraphics[width=\linewidth]{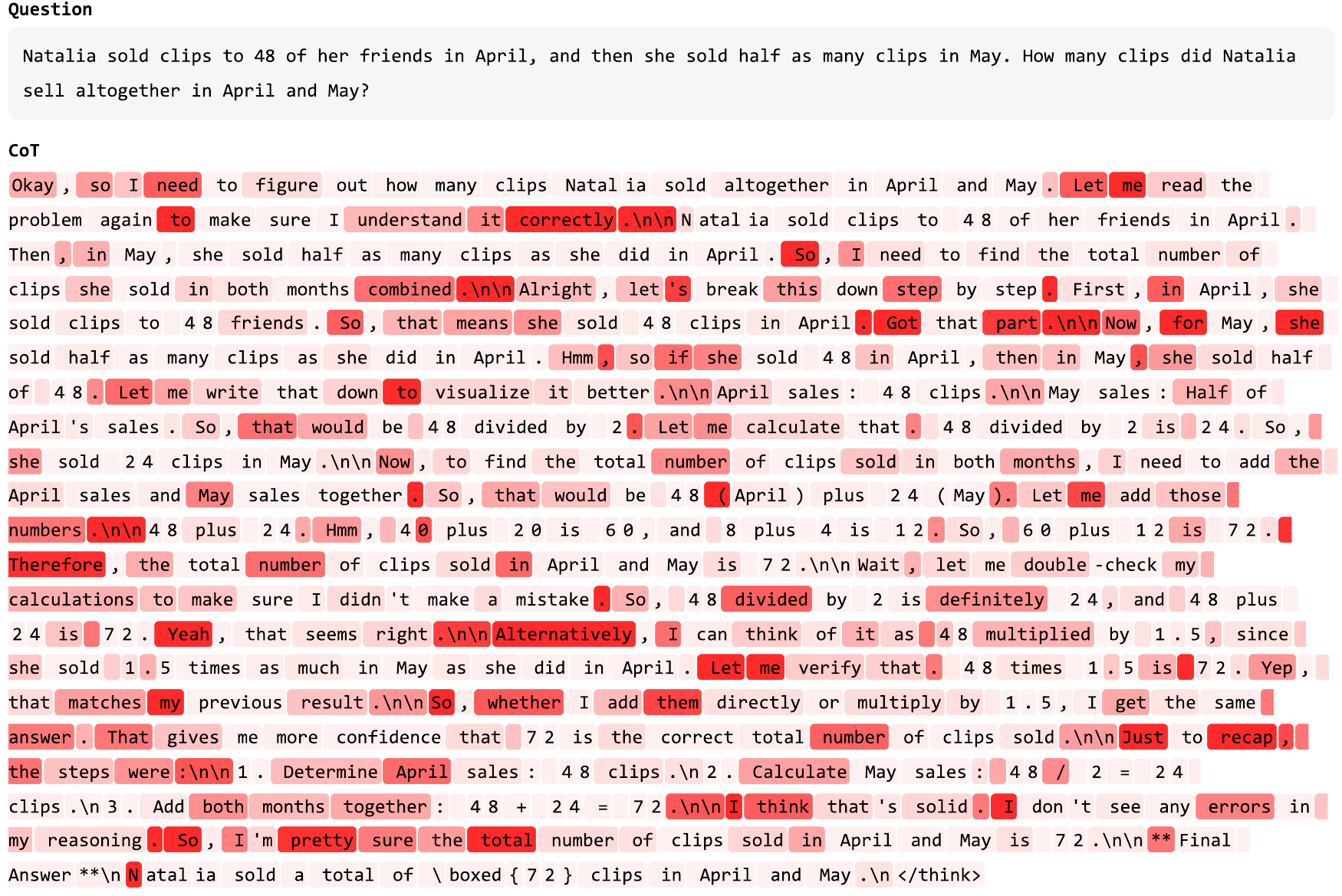}
    \caption{Token-level KL divergence between the original model and the on-policy SFT–trained efficient model along the generated CoT for a GSM8K example. Darker colors indicate larger divergence (Example 1).}
    \label{fig:kl_example1}
\end{figure}

\begin{figure}[t]
    \centering
    \includegraphics[width=\linewidth]{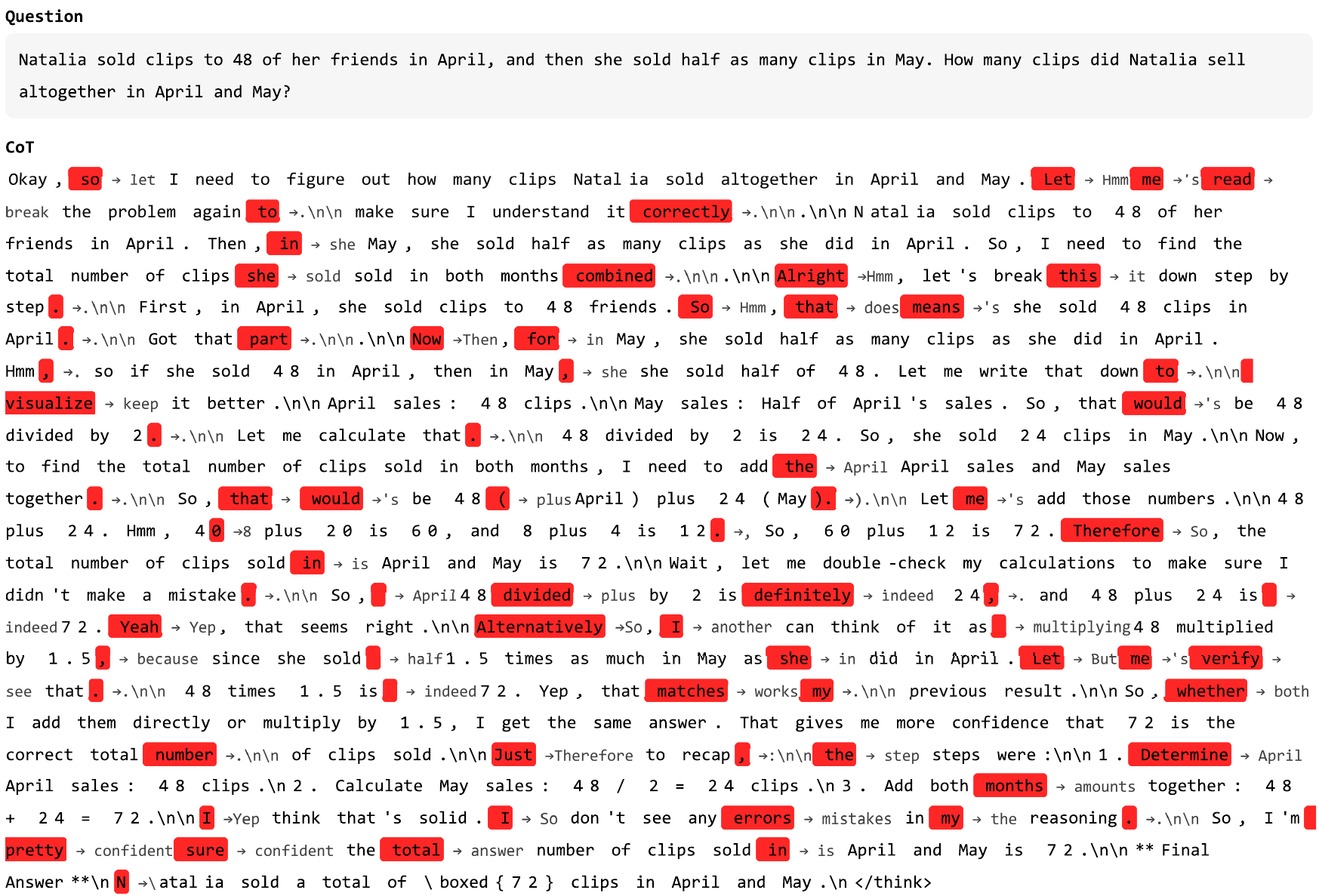}
    \caption{Tokens with the largest KL divergence and the corresponding alternative tokens selected by the efficient model for a GSM8K example (Example 1).}
    \label{fig:mismatch_example1}
\end{figure}

\begin{figure}[t]
    \centering
    \includegraphics[width=\linewidth]{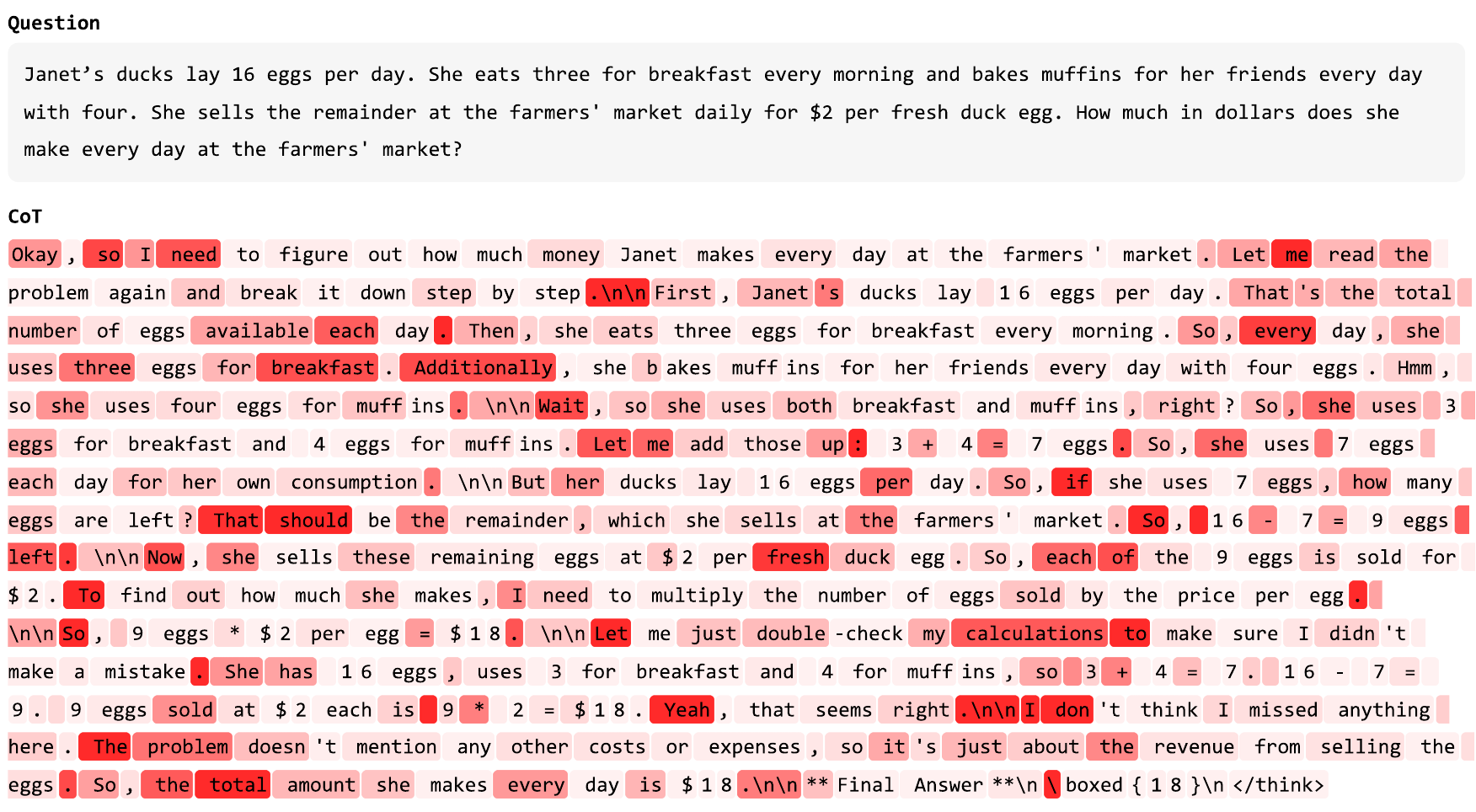}
    \caption{Token-level KL divergence between the original model and the on-policy SFT–trained efficient model along the generated CoT for a GSM8K example. Darker colors indicate larger divergence (Example 2).}
    \label{fig:kl_example2}
\end{figure}

\begin{figure}[t]
    \centering
    \includegraphics[width=\linewidth]{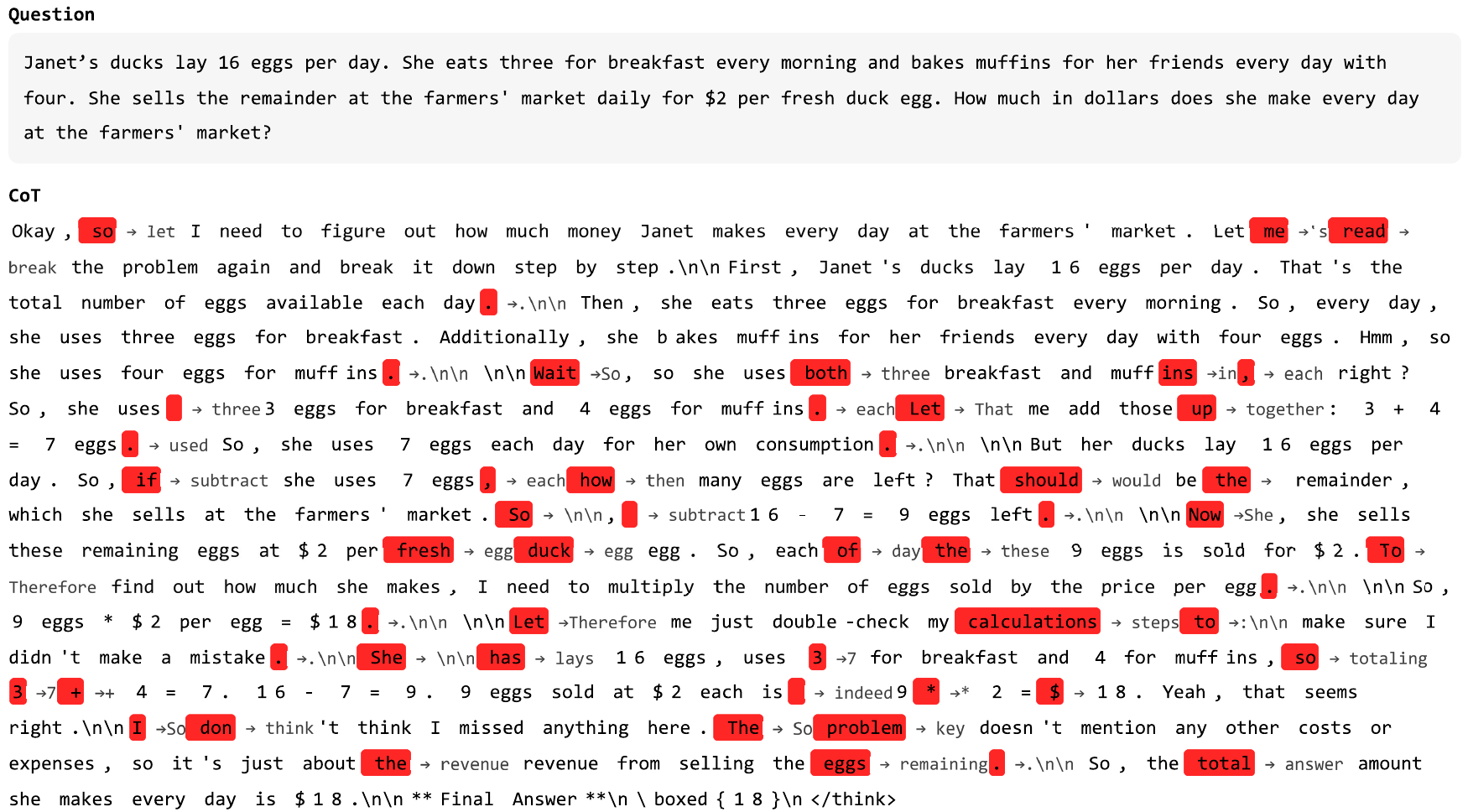}
    \caption{Tokens with the largest KL divergence and the corresponding alternative tokens selected by the efficient model for a GSM8K example (Example 2).}
    \label{fig:mismatch_example2}
\end{figure}

In this section, we present qualitative, question-level examples to illustrate how the proposed diagnostic identifies tokens associated with inefficient reasoning. We provide two concrete examples, each consisting of two complementary visualizations.
For each question, we first visualize the token-level KL divergence between the original model and its on-policy SFT–trained efficient counterpart along the generated CoT. The divergence is computed at each token position under a fixed prefix, with larger values highlighted in darker red, indicating stronger disagreement between the two models. The resulting visualizations are shown in \autoref{fig:kl_example1} and \autoref{fig:kl_example2}.
Second, we capture a subset of tokens with the largest KL divergence and display the alternative tokens selected by the efficient model at the same positions, as shown in \autoref{fig:mismatch_example1} and \autoref{fig:mismatch_example2}.

From the KL visualizations, we observe that tokens such as ``Wait,'' ``Alternatively,'' and ``Hmm,'' which often signal hesitation or shifts in reasoning, consistently exhibit larger divergence. From the alternative-token analysis, we further find that the efficient counterpart frequently replaces tokens such as ``.'' with ``.\textbackslash n\textbackslash n'', suggesting a tendency to terminate the current line of analysis and transition more decisively to the next reasoning step.
Together, these case studies demonstrate that our proposed diagnostic provides a reliable mechanism for localizing inefficiency-inducing token positions, paving the way for more interpretable analyses of efficiency and future research on the development of efficient reasoning models.

\end{document}